\documentclass[final]{iccv}

\usepackage{times}
\usepackage{epsfig}
\usepackage{graphicx}
\usepackage{amsmath}
\usepackage{amssymb}
\usepackage{multirow}
\usepackage{pifont}
\usepackage{booktabs}
\usepackage{epsfig}
\usepackage{booktabs,multirow}
\usepackage{graphics}
\usepackage{threeparttable}
\usepackage{color}
\usepackage[normalem]{ulem}
\usepackage{multirow}
\usepackage{float}
\usepackage{amsfonts}
\usepackage{bm}
\usepackage{rotating}
\usepackage{enumitem}
\usepackage[numbers]{natbib}
\usepackage{array}
\usepackage[table]{xcolor}
\usepackage{colortbl}
\definecolor{myy}{RGB}{126,95,0}
\definecolor{mygray}{gray}{.9}
\definecolor{bblue}{RGB}{30,80,120}
\definecolor{mygray1}{gray}{.7}
\usepackage{bm}
\usepackage{mathtools}
\usepackage{subcaption}
\usepackage{siunitx}
\usepackage[nopar]{lipsum}
\usepackage[export]{adjustbox}

\newcolumntype{I}{!{\vrule width 1pt}}

\definecolor{ggray}{RGB}{127,127,127}
\definecolor{mygreen}{RGB}{93,174,86}
\definecolor{myred}{RGB}{192,0,0}

\makeatletter
\newcommand{\thickhline}{%
	\noalign {\ifnum 0=`}\fi \hrule height 1pt
	\futurelet \reserved@a \@xhline
}
\makeatother

\usepackage{caption}
\usepackage[pagebackref=true,breaklinks=true,colorlinks,bookmarks=false]{hyperref}
\usepackage[utf8]{inputenc}

\usepackage{cleveref}
\crefname{section}{§}{§§}
\Crefname{section}{§}{§§}

\def\iccvPaperID{6126} 
\def\confYear{ICCV 2021}
\begin{document}


\title{Collaborative Visual Navigation}

	%
\author{Haiyang Wang$^{1,2}$~,~~\hspace{1pt}Wenguan Wang$^{3}\thanks{Corresponding author: Wenguan Wang.}$~,~~Xizhou Zhu$^{2}$, ~~Jifeng Dai$^{2}$, ~~Liwei Wang$^{1}$   \\
     {\small {$^1$}  Key Laboratory of Machine Perception, MOE, Peking University \hspace{0pt}} 
     {\small{$^2$}  SenseTime Research \hspace{0pt}
     {$^3$}  Computer Vision Lab, ETH Zurich \hspace{0pt}}\\
     \small wanghaiyang@stu.pku.edu.cn, wenguanwang.ai@gmail.com,  \{daijifeng, zhuwalter\}@sensetime.com, wanglw@cis.pku.edu.cn
     \\ \small\url{https://github.com/Haiyang-W/MAVN}
	%
}

\maketitle


%
%

\begin{abstract}
As a fundamental problem for Artificial Intelligence, multi-agent system (MAS) is making rapid progress, mainly driven by multi-agent reinforcement learning (MARL) techniques. However, previous MARL methods largely focused on grid-world like or game environments; MAS in visually-rich environments has remained less explored. To narrow this gap and emphasize the crucial role of perception in MAS, we propose a large-scale 3D dataset, CollaVN, for multi-agent visual navigation (MAVN). In CollaVN, multiple agents are entailed to cooperatively navigate across photo-realistic environments to reach target locations. Diverse MAVN variants are explored to make our problem more general.  Moreover, a memory-augmented communication framework is proposed. Each agent is equipped with a private, external memory to persistently store communication information. This allows agents to make better use of their past communication information, enabling more efficient collaboration and robust long-term planning.  In our experiments, several baselines and evaluation metrics are designed. We also empirically verify the efficacy of our proposed MARL approach across different MAVN task settings. 
\end{abstract}

\section{Introduction}
Human intelligence would collectively solve the problem that otherwise is unthinkable by a single person. For instance, Condorect's Jury Theorem~\cite{landemore2008democratic} indicates that, under certain assumption, adding more voters in a group would increase the probability that the majority chooses the right answer. Thus it is believed that a next grand challenge of AI is to answer how multiple intelligent agents could learn human-level collaborations~\cite{goertzel2007artificial,peng2017multiagent}. With the recent rapid advances of machine learning and robotic technologies, the collaboration of AI agents gathers significantly increasing attention from researchers, and finds great potential in a wide range of real-world applications, such as search-and rescue~\cite{liu2016multirobot}, area coverage~\cite{stergiopoulos2015distributed, kantaros2015distributed}, harbour protection~\cite{miah2014nonuniform}, and so on. In short, such multi-agent system (MAS)~\cite{stone2000multiagent} is of central importance to AI and still in its infancy.

\begin{figure}[t]
\vspace{-4pt}
  \centering
      \includegraphics[width=0.99\linewidth]{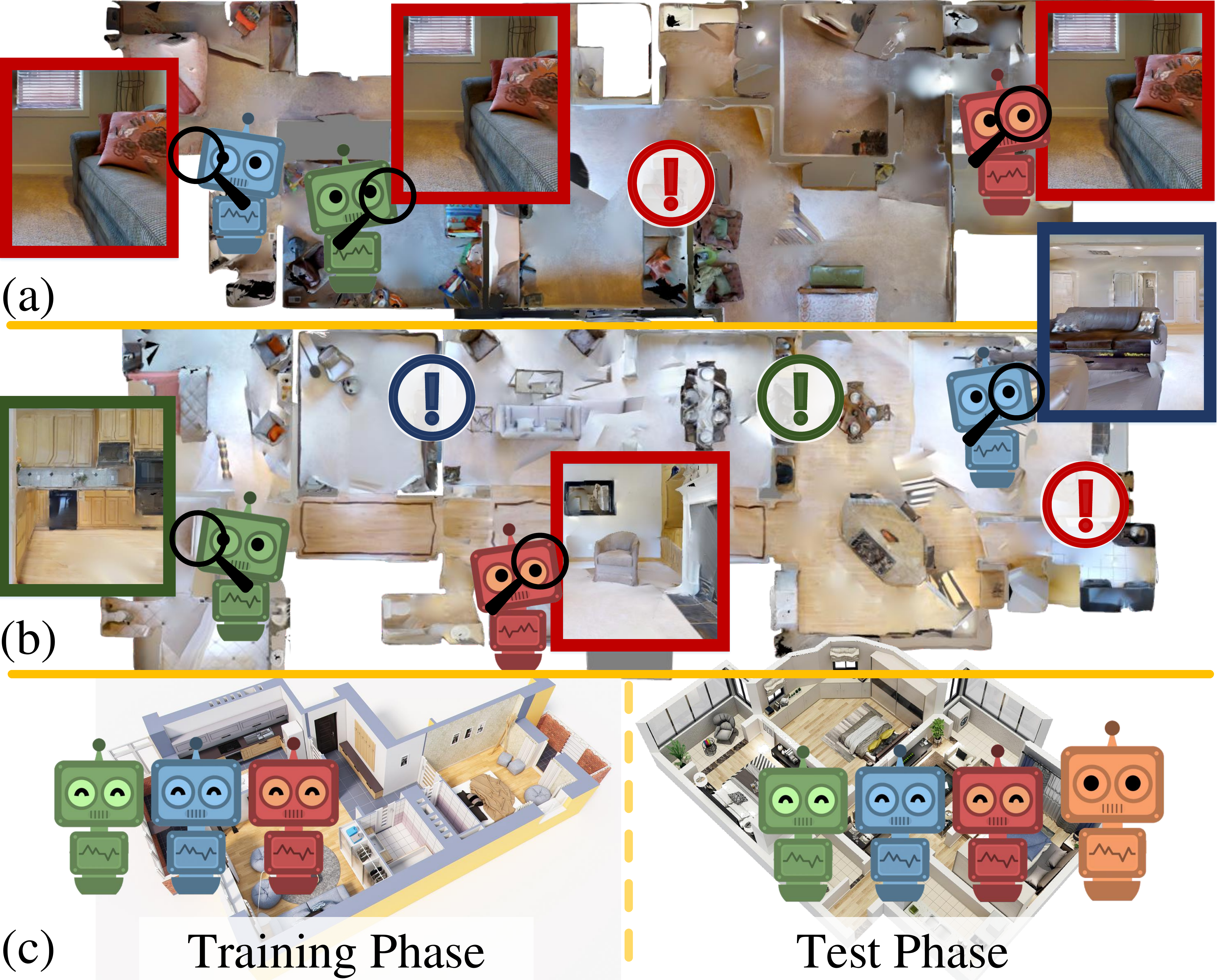}
\captionsetup{font=small}
	\caption{\small{~\!Three collaborative visual navigation task settings in CollaVN: (a)~\!CommonGoal, (b)~\!SpecificGoal, and (c)~\!Ad-hoCoop.
}
}
\label{fig:overview}
\end{figure}



Multi-agent reinforcement learning (MARL)\!~\cite{busoniu2008comprehensive} and embodied systems\!~\cite{Peng_2018} are the main driving power behind MAS. MARL field is rapidly expanding, and some of approaches even show super-human level performance in various games~\cite{jaderberg2019human}, such as Magent\!~\cite{zheng2017magent}, StarCraft\!~\cite{samvelyan2019starcraft}, and Dota2~\cite{openai2019dota}. However, most of existing algorithms~\cite{lowe2017multi, rashid2018qmix, foerster2016learning, lazaridou2017multiagent} operate on  low-dimensional observations, \eg, grid-world like or game environments.  As for embodied systems, the availability of large-scale, visually-rich 3D datasets~\cite{armeni20163d, chang2017matterport3d, wu2018building,song2017semantic} and community interest in active perception led to a recent surge of simulation platforms~\cite{shen2020igibson, chang2017matterport3d, kolve2017ai2, savva2019habitat, xia2018gibson}. Based on these high-performance 3D simulators, various active vision tasks are investigated, such as embodied navigation~\cite{chaplot2020neural,deitke2020robothor}, instruction following~\cite{anderson2018vision}, embodied question-answering~\cite{das2018embodied} and perception~\cite{liu2020when2com}, \etc. However, most of them either do not support multi-agent setting~\cite{chaplot2020neural,deitke2020robothor,anderson2018vision,das2018embodied}, or little consider planning~\cite{das2018embodied,liu2020when2com}.

Thus there is still a huge gap between MAS and realistic visual environments. To close this gap, we introduce a collaborative visual navigation (MAVN) task for image-goal based, multi-agent navigation, under visually realistic environments, and develop a corresponding dataset, called CollaVN. Autonomous navigation forms a core building block in embodied systems, and long received research interest across many fields, such as computer vision~\cite{chaplot2020neural}, robotics~\cite{kim1999symbolic},  linguistics\!~\cite{striegnitz2011report}, and cognition\!~\cite{kuipers1978modeling}. In MAVN, agents cooperate to reach a target position (determined by a corresponding image), with their individual perception. Three task settings are explored to cover more challenges in MAS/MARL: i) \textbf{{CommonGoal}} (Fig.~\ref{fig:overview}(a)), where agents are assigned \textit{a same goal picture}; ii) \textbf{{SpecificGoal}} (Fig.~\ref{fig:overview}(b)), where agents are assigned \textit{different goal pictures;} and iii) \textbf{{Ad-hoCoop}} (Fig.~\ref{fig:overview}(c)), where the \textit{agent numbers} in training and testing are different. \textit{CommonGoal} serves as a basic setting where agents learn how to cooperate to reach a goal faster. \textit{SpecificGoal} captures more complex scenarios where agents need to collaborate with individual targets.  \textit{Ad-hoCoop} is built upon \textit{CommonGoal} but agents are required to adapt to different team sizes during deployment, simulating a more open MAS situation~\cite{barrett2011empirical,canaan2019diverse}.

Our CollaVN dataset is built upon iGibsonV1 simulator~\cite{xia2020interactive}. Besides inheriting the advantages of iGibsonV1 in high-quality rendering, rich diversity and good accessibility, CollaVN has the following unique characteristics:
\begin{itemize}[leftmargin=*]
	\setlength{\itemsep}{0pt}
	\setlength{\parsep}{-2pt}
	\setlength{\parskip}{-0pt}
	\setlength{\leftmargin}{-10pt}
	\vspace{-6pt}
	\item \textbf{Multi-agent operation:} CollaVN supports MAVN with arbitrary group sizes (2-4 agents used in current setting). Agents can be safely initialized with collision detection.
    \item \textbf{Multi-task setup:}  All the three task variants are involved and companied with standard evaluation tools.
	\item \textbf{Flexible target generation:} CollaVN allows to generate different target pictures flexibly, well supporting the three MAVN settings and future dataset extension.
    \item \textbf{Panoramic$_{\!}$ observation:}$_{\!}$ A$_{\!}$ panoramic$_{\!}$ camera$_{\!}$ is$_{\!}$ equipped$_{\!}$ for rendering agent  egocentric panorama in real-time.
	\vspace{-4pt}
\end{itemize}

We further propose a novel memory-augmented communication approach for MAVN. Communication is viewed as one of the most desired and efficient ways of multi-agent interaction/coordination~\cite{kim2019learning,lazaridou2017multiagent}. Communication is especially crucial when a set of
cooperative agents are placed in a partially observable environment, such as our MAVN setting -- the agents need to exchange information (\eg, past experience, current perception, future plan, \etc), coordinate their actions, behave as a group and achieve the target goal better~\cite{fox2000probabilistic,Hernandez_Leal_2019}. Thus learning communication has become a center topic in MARL field. From earlier natural-language-based~\cite{lazaridou2017multiagent} to recent hidden-state-based~\cite{peng2017multiagent} communication forms and pre-defined communication protocols~\cite{tan1993multi} to learnable strategies~\cite{foerster2016learning,peng2017multiagent,sukhbaatar2016learning}, communication based MARL made great advance. Some most recent ones are concerned with practical communication-bandwidth-limited situations~\cite{singh2018learning,das2019tarmac,liu2020when2com,liu2020who2com} and conduct studies mainly around the theme of learning \textit{what}, \textit{when} and \textit{who} to communicate. However, in previous methods, agents simply discard all the information after each communication round. This is problematic, as some communication information, even not useful in present, may be valuable in the future. This raises the risk of losing essential information and easily leads to suboptimal behaviors driven by short-termism. To address these limitations, in our memory-augmented communication approach, each agent is equipped with a private, compartmentalized memory, for safely storing and accurately recalling its past communication information. During each communication round, each agent can send a more reasonable request to other agents, by considering its current states and stored information. Each of other agents will check its current state and past communication information to give a better response. The agents can make full use of the rich information stored in the memory during navigation. Overall, our approach helps agents conduct more efficient communication and enhances their long-term planning ability, eventually leading to better collaboration and navigation performance.

In a nutshell, our contributions are three-fold:
\begin{itemize}[leftmargin=*]
	\setlength{\itemsep}{0pt}
	\setlength{\parsep}{-2pt}
	\setlength{\parskip}{-0pt}
	\setlength{\leftmargin}{-10pt}
	\vspace{-6pt}
    \item To narrow the gap between MAS and visual perception, we develop a large-scale and publicly-accessible dataset, CollaVN, for collaborative visual navigation (MAVN), in photo-realistic, multi-agent environments.
    \item Diverse MAVN task settings are explored to cover many core challenges in MAS/MARL, with packaged evaluation tools and baselines.
    \item A novel memory-augmented communication approach is proposed to address more efficient multi-agent collaborations and robust long-term planning.
	\vspace{-4pt}
\end{itemize}

Several baselines are evaluation tools are developed for comprehensive benchmarking. Experimental results show that our memory-augmented communication framework achieves promising performance over the three MAVN settings on CollaVN dataset.  This work is expected to provide insights into the critical role of MAS in embodied perception tasks, and foster research on the open issues raised.

\section{Related Work}%


\noindent\textbf{Vision-based Navigation.} As a crucial step towards building intelligent robots,  autonomous navigation has been long studied in robotics community. Recently vision-based navigation~\cite{chaplot2020neural} attained growing attention in computer vision community and was explored in many other forms, such as point-goal based or object-oriented (\ie, reaching a specific position or finding an object)~\cite{zhu2016targetdriven,chaplot2020neural, chaplot2020object,deitke2020robothor}, natural-language-guided~\cite{anderson2018vision,wang2020active}, audio-assisted navigation~\cite{chen2020soundspaces}, and from indoor environments~\cite{anderson2018vision} to street scenes~\cite{chen2019touchdown}. Prominent early methods rely on a pre-given/-computed global map~\cite{borenstein1991vector,kim1999symbolic} for path planing, while later ones refer to simultaneous localization and mapping (SLAM) techniques~\cite{thrun2002probabilistic,hartley2003multiple} that reconstruct map on the fly. Some more recent methods learn planning and mapping jointly~\cite{gupta2017cognitive,parisotto2017neural,zhang2017neural,chaplot2020learning,lee2018gated,Wang_2021_CVPR}.

Although great advance has been achieved, visual navigation in multi-agent setting has remained largely unexamined. Most existing studies~\cite{anderson2018vision,chen2020soundspaces,anderson2018vision,chen2019touchdown} focused on single-agent navigation, excepting \cite{jain2020cordial,jain2019two} considering two-agent collaboration in finding and lifting bulky items. In this work, we develop a large-scale dataset for indoor, collaborative, multi-agent visual navigation. The agents are required to work together to perform same or different tasks, or even without any assumption on the number of involved agents. Thus our task setting is more novel and general.

\noindent\textbf{Embodied AI Environments.}  The recent surge of research in embodied perception is greatly driven by new 3D environments~\cite{xia2018gibson,armeni20163d, chang2017matterport3d,brodeur2017home,song2017semantic} and simulation platforms, such as iGibsonV1~\cite{xia2020interactive}, AI2-THOR~\cite{kolve2017ai2}, and Habitat~\cite{savva2019habitat}.
Compared with grid-like or game environments~\cite{zheng2017magent,samvelyan2019starcraft,openai2019dota,jaderberg2019human},  these open accessible, photo-realistic platforms bring perception and planning in a close loop and make the training and testing of embodied agents reproducible~\cite{chen2020soundspaces}. Based on these platforms, numerous embodied vision datasets were proposed~\cite{deitke2020robothor,anderson2018vision,chen2020soundspaces,deitke2020robothor}, while most of them are designed for single-agent tasks. We build our dataset upon iGibsonV1~\cite{xia2020interactive} which is featured by large scale, rich visual diversity and strong extensibility. Although iGibsonV2~\cite{shen2020igibson} also involves multiple agents, it addresses obstacle avoidance in social scenes. It does not investigate collaboration among agents, while which is the core nature and research focus in MAS/MARL. Moreover, our dataset supports diverse essential MARL task settings, making it unique in the filed of visual navigation.

\noindent\textbf{Multi-Agent Reinforcement Learning (MARL).} MARL tackles$_{\!}$ the$_{\!}$ sequential$_{\!}$ decision-making$_{\!}$ problem$_{\!}$ of$_{\!}$ multiple$_{\!}$ agents in a common environment, each of which aims to achieve its own long-term goal by interacting with the environment and other agents~\cite{busoniu2008comprehensive,Hernandez_Leal_2019}. Basically, MARL algorithms can be placed into four categories~\cite{Hernandez_Leal_2019}. i) \textit{Analysis of emergent behaviors}. Some early studies~\cite{raghu2018can, bansal2018emergent} analyze single-agent RL algorithms in multi-agent environments. ii) \textit{Learning communication}. Methods in this category~\cite{mordatch2018emergence,lazaridou2017multiagent, foerster2016learning} address collaboration through explicit communication, which attracts increasing attention recently. iii) \textit{Learning cooperation}. Many other efforts~\cite{panait2005cooperative, matignon2012independent, palmer2017lenient} indirectly arrive at cooperation via, for example, policy parameter sharing~\cite{gupta2017cooperative} or experience replay buffer~\cite{foerster2017stabilising}. iv) \textit{Agents modeling agents.} These methods~\cite{raileanu2018modeling, hong2017deep} build models to reason the behaviors of other agents for better decision-making.

These algorithms mostly focus on grid-world like scenes, despite the role of perception to some degree. Our CollaVN dataset provides a visually-rich platform for fostering studies about MARL in computer vision. We consider MAVN is of a cooperative nature and agents are situated in a partially observable environment, and concern with learning efficient communication for long-term and collaborative planning.

\noindent\textbf{Learn to Communicate.} Communication is a fundamental aspect of intelligence, enabling agents to behave as a group, rather than a collection of individuals. Along with the direction of communication based collaboration, MARL researchers first used predefined communication protocols~\cite{tan1993multi}, and then learnable strategies~\cite{foerster2016learning,peng2017multiagent,jiang2018learning,sukhbaatar2016learning} for information exchanging. However, sharing information among all agents is problematic, as communication bandwidth is limited the real world. To reduce bandwidth wasting, some recent methods adopt pre-defined communication groups~\cite{jiang2018learning,Jiang2020Graph} or set ``what, when and/or who to communicate'' as a part of policy learning~\cite{foerster2016learning,singh2018learning,das2019tarmac,liu2020when2com,liu2020who2com}.

As previous methods only have an ``immediate memory'' during communication, they suffer from a bias towards short-sighted behaviors. We instead equip each agent with a private, external memory that persistently stores the information in all past communication rounds and enables the reuse of these information during future navigation. This enhances the long-term planning ability of our agents. This idea is also distinctively different from~\cite{peng2017multiagent,pesce2020improving}, where all the agents need to transmit messages to a central memory, incurring a major communication bottleneck.

\noindent\textbf{Ad-Hoc Cooperation.} The problem of ad-hoc team play in multi-agent cooperative games was raised in the early 2000s~\cite{bowling2005coordination} and is mostly studied in the robotic soccer domain~\cite{hausknecht2016half}. In many close MARL environments agents learn a fixed and team-dependent policy~\cite{zhang2020multi}. While in ad-hoc setting agents are desired to assess and adapt to the capabilities of others to behave optimally in an open environment. See~\cite{albrecht2018autonomous} for a survey. Existing work in ad-hoc MARL mainly focuses on game settings, such as hanabi (card game)~\cite{canaan2019diverse}. Our work makes a further step towards exploring ad-hoc setting in visually-rich MARL; the navigation agents are required to adapt to different team sizes during deployment.


\section{Multi-Agent Visual Navigation Task}

\subsection{Our CollaVN Dataset}\label{sec:3.1}


CollaVN is a visual MARL dataset, created for the development of intelligent, cooperative navigation agents.

\noindent\textbf{iGibsionV1~\cite{xia2020interactive}.} CollaVN is built upon iGibsonV1, a recently released open-source 3D simulator.  iGibsonV1 supports fast visual rendering in 400 fps and  allows researchers to easily train and evaluate embodied navigation agents.


\noindent\textbf{New Modules.} To support multi-agent navigation task, three modules are developed: i) An agent initialization module is responsible for ``safely'' placing multiple agents at the beginning of each navigation episode, \ie, randomly placing agents but being aware of constraints of scene and physics, such as avoiding collision. ii) As the panoramic visual observation space is widely adopted in current embodied vision tasks~\cite{chaplot2020neural, deng2020evolving}, a panoramic camera is equipped to render agents 360$^\circ$ egocentric views in real-time. iii) A target generation module is built for generating a target picture as the goal of navigation. This module adopts a front-view camera, taking pictures around the agent height (\ie, 0.9 m) under the constraints of scene and physics.

\noindent\textbf{Scenes.} All the 572 full buildings in iGibsonV1 are involved in CollaVN, covering a total area of 211K $\text{m}^{2}$. The area of a physical navigation space considered in CollaVN is from 10 m $\times$ 10 m to 30 m $\times$ 30 m.

\noindent\textbf{Agent.} In CollaVN, we use Locobot, a widely used simulation agent~\cite{pyrobot2019}. Its body width is 0.36 m, controlled by two wheels. The maximum translational velocity is 70 cm/s and rotational velocity is 180 deg/s.

\noindent\textbf{Data Generation.} Following Gibson-tiny split~\cite{xia2020interactive},  we use 25/5/5 scenes for creating \texttt{train}/\texttt{val}/\texttt{test} data. We build three sub-dasets for different MAVN tasks, \ie, \textit{CommonGoal}, \textit{SpecificGoal}, \textit{Ad-hoCoop}. A total of 1M/60K/120K episodes/samples are generated for \texttt{train}/ \texttt{val}/\texttt{test}. According  to initial agent-target distance at the beginning of each episode, fine-grained annotations for \texttt{val}/ \texttt{test} data are given: $\{$\textit{easy} (1.5-3 m),  \textit{medium} (3-5~m), \textit{hard} (5-10 m)$\}$. We provide more details in \S\ref{sec:ts}.

\noindent\textbf{Comparison$_{\!}$ to$_{\!}$ Previous$_{\!}$ Datasets.}$_{\!}$ As$_{\!}$ shown$_{\!}$ in$_{\!}$ Table~\ref{tab:comparison},$_{\!}$ Habitat~\cite{savva2019habitat} only supports the single-agent setting. AI2THOR~\cite{kolve2017ai2} just supported multi-agent setting recently. However, its main focus is task planning of sequential actions to change object states and
its environment space is small (only 9 m $\times$ 4 m), making it not suitable for studying challenging, long-range navigation. Note that \cite{jain2020cordial,jain2019two} recently extended AI2THOR to involve a collaborative task, \ie, finding and lifting furniture. They are limited to highly-correlated actions between two agents and cannot explore complex multi-agent collaborative behaviors under more general situations (\eg, larger team sizes, different goals). iGibsonV2~\cite{shen2020igibson}, also built upon iGibsonV1, mainly focuses on obstacle avoidance in social scenes, \ie, only one agent is required to execute point-goal navigation while other agents are moving around (acting as pedestrians). Thus iGibsonV2 does not address collaboration among multiple navigators.

\begin{table}
	\centering
	\small
	\resizebox{0.49\textwidth}{!}{
		\setlength\tabcolsep{6pt}
		\renewcommand\arraystretch{1.0}
		\begin{tabular}{l|c|c||ccccc}
			\hline\thickhline
			\rowcolor{mygray}
Dataset &Pub. &Year & MA  & CM & LS & CA & PA \\\hline\hline
AI2THOR~\cite{kolve2017ai2}  &arXiv &2017 & \checkmark  & \checkmark  &  &  & \\
Habitat~\cite{savva2019habitat} &ICCV&2019 &   &   & \checkmark &  & \checkmark\\
iGibsonV1~\cite{xia2020interactive} &RAL&2020  &   &  & \checkmark & \checkmark &  \\
iGibsonV2~\cite{shen2020igibson} &arXiv&2020 & \checkmark  &  & \checkmark & \checkmark &  \\
\hline
\textbf{CollaVN} &-&2021 & \checkmark  & \checkmark  & \checkmark & \checkmark & \checkmark \\\hline
		\end{tabular}
	}
	\captionsetup{font=small}
	\caption{\small Point-to-point comparison to closely relevant 3D platforms/datasets. MA: Multi-agent. CM: Multi-agent communication. LS: Large scale. CA: Continuous action. PA: Panorama.}
	\label{tab:comparison}
\end{table}

\subsection{Task Setup}\label{sec:ts}

\noindent\textbf{Basic Setting.} MAVN assumes there are $N$ autonomous agents situated in a CollaVN environment. Each agent is required to navigate the environment to reach a target location (indicated by a goal image), under a cooperative setting. Formally, at the beginning of each navigation episode, each agent $n$ will be assigned a target goal image $g_{n}$, \ie, a $3\!\times\!128\!\times\!128$ RGB image. Each agent does not know other agents' positions. At each time step $t$, each agent $n$ receives its visual observation $O^{t}_n$, \ie, a first-person panoramic-view $3\!\times\!512\!\times\!128$ RGB image.  As in many communication based MARL settings~\cite{jiang2018learning, foerster2016learning, kim2019learning}, agents operate in a partially observable environment and perceive the environment from their own view. They can exchange information via a low bandwidth communication channel. Sensing and control errors and communication delay are not considered.

\noindent\textbf{Action Space.} The agents can control its wheels, and the maximum allowed translational and rotational velocities are 70 cm/s and 180 deg/s, respectively. At each time step $t$, each agent $n$ takes a continuous action $a^{t}_n\!\in\![-1, 1]^{2}$, corresponding to the normalized velocities. For example, $a^{t}_n\!=\!(1, 0.5)$ means the agent will move at the maximum translational velocity (\ie, 70 cm/s) and half of the maximum rotational velocity (\ie, 90 deg/s) during [$t$, $t\!+\!1$]. The interval between $t$ and $t\!+\!1$ is 1 s.

\noindent\textbf{MAVN Task Settings.}  To better cover challenges in MARL and MAS, three different MAVN task settings are designed:
\begin{itemize}[leftmargin=*]
	\setlength{\itemsep}{0pt}
	\setlength{\parsep}{-2pt}
	\setlength{\parskip}{-0pt}
	\setlength{\leftmargin}{-10pt}
	\vspace{-6pt}
	\item \textbf{CommonGoal:} $N$ robot agents are assigned a common goal, \ie, $g_{1}\!=\!\cdots\!=\!g_{n}\!=\!\cdots\!=\!g_{N}$. Agents need to collaboratively navigate to the target goal within a maximum time step length $T$. This is the most basic MAVN task setting with the purpose of investing how several agents can learn, from visual environment, to communicate so as to effectively and collaboratively solve a same given task.
    \item \textbf{SpecificGoal:}  The difference from \textit{CommonGoal} is that, \textit{SpecificGoal} assigns $N$ agents with $N$ different goal images, \ie, $g_{1}\!\neq\!\cdots\!\neq\!g_{n}\!\neq\!\cdots\!\neq\!g_{N}$. In this setting, we are particularly interested in how agents learn communication and collaboration with different long-term targets.
	\item \textbf{Ad-hoCoop:} In \textit{CommonGoal} and \textit{SpecificGoal}, the team size remains fixed during both training and deployment phases. The learned agents may not generalize to new configurations of teams. \textit{Ad-hoCoop} is a more open world setting, in which agents must adapt to different team sizes. We test the ad-hoc team play by adding or removing agents at test time. Note that in  \textit{Ad-hoCoop}  agents are assigned a same goal in each   episode.
\end{itemize}

\noindent\textbf{Task-Specific Sub-Dataset.} Our CollaVN has three sub-datasets, correspond to the three MAVN tasks:
\begin{itemize}[leftmargin=*]
	\setlength{\itemsep}{0pt}
	\setlength{\parsep}{-2pt}
	\setlength{\parskip}{-0pt}
	\setlength{\leftmargin}{-10pt}
	\vspace{-6pt}
	\item \textbf{CommonGoal-CollaVN}: Three subsets are created for different agent numbers (\ie, $N\!=\!\{2,3,4\}\!$) and agents in each episode are assigned a same target image. For each subset, we generate 5K episodes per training scene. For each val/test scene, we generate 0.5K/1K episodes per configuration, \ie, $\{$\textit{easy} (1.5-3 m),  {\textit{medium} (3-5 m)}, {\textit{hard} (5-10 m)}$\}$.  Finally, each subset has 125K/7.5K/15K samples for \texttt{train}/\texttt{val}/\texttt{test}.

\item \textbf{SpecificGoal-CollaVN}: Three subsets are created for different agent numbers (\ie, $N\!=\!\{2,3,4\}\!$) and agents in each episode are assigned different target images.
Similarly, each subset has 125K/7.5K/15K samples for \texttt{train}/\texttt{val}/ \texttt{test}.

\item \textbf{Ad-Hoc-CollaVN}: Two subsets are created for different ad-hoc team size changing situations (\ie, $N\!\!:\!2\!\rightarrow\!3$ and $N\!\!:\!3\!\rightarrow\!2$), and each has 125K/7.5K/15K samples for \texttt{train}/\texttt{val}/\texttt{test}. Agents in each episode are assigned a same target image.  For $N\!\!:\!2\!\rightarrow\!3$, its \texttt{train} set is the one in  \textit{CommonGoal}-CollaVN with $N\!=\!2$, but its \texttt{val} and \texttt{test} sets are new generated. Each training sample in $N\!=\!2$ contains 2 agents while each val/test sample has 3 agents. The subset of $N\!\!:\!2\!\rightarrow\!3$ is built in a similar way.
\end{itemize}


%


\subsection{Evaluation Measures}\label{sec:3.3}
For performance evaluation, we use four metrics, where the former three are multi-agent augmented versions of SR (Success Rate), DTS (Distance to Success) and SPL (Success weighted by Path Length~\cite{anderson2018evaluation}), and the last one, named SSR (Success weighted by Step Ratio), is new proposed.

\noindent\textbf{SR.} Following~\cite{xia2020interactive}, we consider an episode is successful for an agent if it reaches the target location within 1 m radius. Let $\tau_{k,n}\!\in\!\{0,1\}$ be a binary indicator of success in episode $k$ for agent $n$. SR is given as: $\frac{1}{KN}\!\sum\nolimits_{k}\!\sum\nolimits_{n\!}\tau_{k,n}$.

\noindent\textbf{DTS.} Let $d^g_{k,n}$ denote the geodesic distance between agent $n$ and its goal location at the end of episode $k$, and $d^s$ the success threshold (\ie, 1 m). $\text{DTS}\!=\!\frac{1}{KN}\!\sum\nolimits_{k\!}\!\sum\nolimits_{n\!}d^g_{k,n\!}\!-\!d^s$.

\noindent\textbf{SPL.} Let $l^g_{k,n}$ be the geodesic distance from starting position to the goal of agent $n$ in episode $k$, and $l_{k,n}$ the length of the path actually taken by agent $n$ in episode $k$. We have:
$\text{SPL}\!=\! \frac{1}{KN}\!\sum\nolimits_{k}\!\sum\nolimits_{n}\tau_{k,n}{l^g_{k,n}}/{\text{max}(l^g_{k,n}, ~l_{k,n})}$.

\noindent\textbf{SSR.} Let $T$ denote the allowed maximum time step and $T_{k,n}$ the number of navigation steps used by agent $n$ in episode $k$. $\text{SSR}\!=\!\frac{1}{KN}\!\sum\nolimits_{k}\!\sum\nolimits_{n\!}\tau_{k,n}{T}/{\text{min}(T,T_{k,n})}$. As MAVN is a very difficult task, in our experiments (\S\ref{section:result}), we found agents usually need to take many steps to reach the targets, \ie, $l^g_{k,n}\!\ll\!l_{k,n}$, making SPL very small and less discriminative. So we design SSR as a complementary.

\section{Methodology}
\begin{figure*}[t]
\vspace{-1pt}
  \centering
      \includegraphics[width=0.99\linewidth]{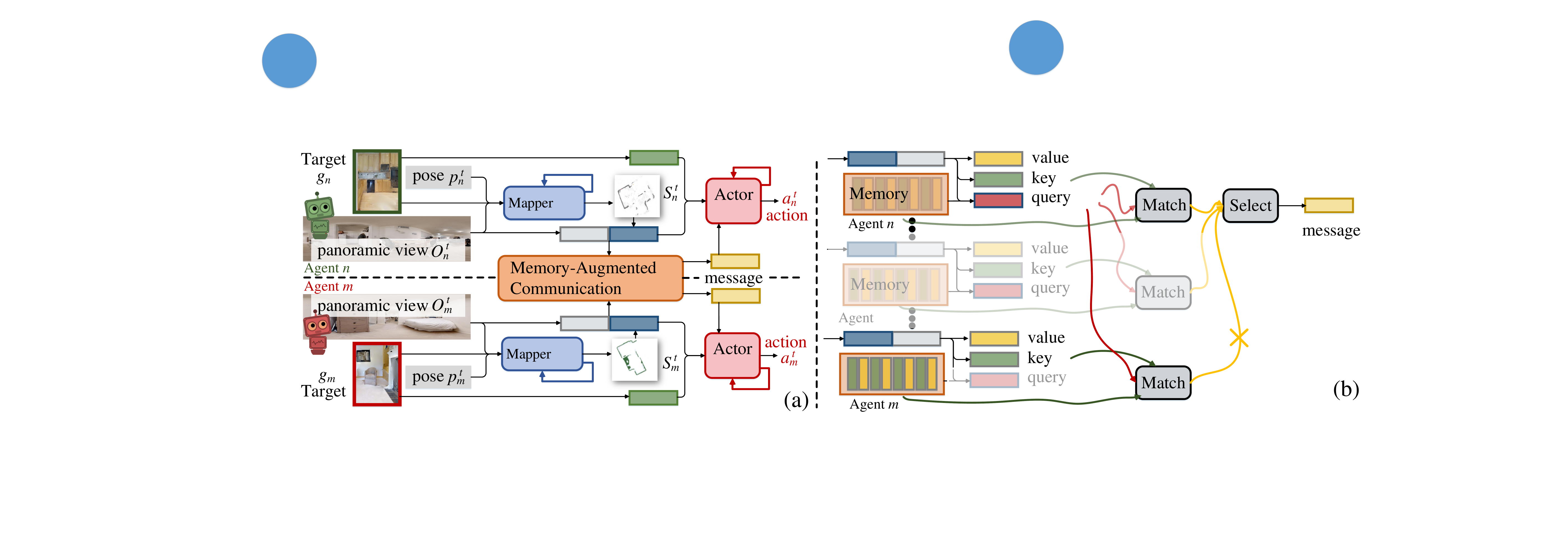}
\vspace{-6pt}
   \put(-376,96){{\tiny $\mathcal{F}^{\text{map}}$}}
   \put(-377,27){{\tiny $\mathcal{F}^{\text{map}}$}}
   \put(-344, 56){{\scriptsize\S\ref{section:com_protocol}}}
   \put(-298, 92){{\scriptsize$\bm{\pi}_{\theta_{n}}$}}
   \put(-298, 22){{\scriptsize$\bm{\pi}_{\theta_{m}}$}}
   \put(-135, 117){{\scriptsize$\bm{\upsilon}^t_n$}}
   \put(-135, 108){{\scriptsize$\bm{\kappa}^t_n$}}
   \put(-135, 99){{\scriptsize$\bm{\mu}^t_n$}}
   \put(-137, 35){{\scriptsize$\bm{\upsilon}^t_m$}}
   \put(-137, 25){{\scriptsize$\bm{\kappa}^t_m$}}
   \put(-278, 69){{\scriptsize$\bm{\omega}^t_n$}}
   \put(-278, 53){{\scriptsize$\bm{\omega}^t_m$}}
   \put(-23, 105){{\scriptsize$\bm{\omega}^t_n$}}
   \put(-60, 20){{\scriptsize$\Gamma(s^t_{n,m}, \eta) = 0$}}
   \put(-211, 99.5){{\scriptsize$\mathcal{M}^t_n$}}
\captionsetup{font=small}
	\caption{\small{(a) Overview of our method. (b) Detailed illustration of our memory-augmented communication strategy.
}
}
\label{fig:framework}
\end{figure*}

\subsection{Problem Setup}
\noindent \textbf{MARL Background.} We consider a system of $N$ agents operating in a common
environment as a $N$-agent extension of partially observable Markov decision processes~\cite{littman1994markov}. It assumes a set~\cite{pesce2020improving}, $\mathcal{S}$, containing all the states characterising the environment; action spaces $\mathcal{A}_1, \mathcal{A}_2, \cdots, \mathcal{A}_N$, where $\mathcal{A}_n$ is a set of possible actions for the $n^{th\!}$ agent; and observation spaces $\mathcal{O}_1, \mathcal{O}_2, \cdots, \mathcal{O}_N$ where $\mathcal{O}_n$ contains the observations available to the $n^{th\!}$ agent. Each agent $n$ receives a private observation $O_{n\!}\!\in\!\mathcal{O}_{n\!}$ (\ie, a partial characterisation of the current state $\mathcal{S}$), and  chooses an action according to a stochastic policy $\bm{\pi}_{\theta_n\!}\!:\! \mathcal{O}_{n\!}\!\mapsto\!\mathcal{A}_n$, parametrized by $\theta_n$. The next state is produced according to a transition function $\mathcal{T}_{\!}\!:\! \mathcal{S}\!\times\!\mathcal{A}_1\!\times\!\cdots\!\times\!\mathcal{A}_{N\!}\!\mapsto\!\mathcal{S}$. Each agent $n$ obtains reward as a function of the state and its action $r_n: \mathcal{S}\!\times\!\mathcal{A}_{n\!}\!\mapsto\!\mathbb{R}$ and aims to maximize its expected return  $R_n\!=\!\sum_{t=1}^T\gamma^tr^t_n(s^t,a_n^t)$, where $\gamma$ is a discount factor and  $T$ is the time horizon.

\noindent \textbf{Task Statement.} With our three MAVN tasks, we explore how agents learn from visually-rich environments to achieve collaborative navigation. In \textit{CommonGoal} (\textit{SpecificGoal}), agents learn to cooperatively navigate to same (different) locations, determined by target images $\{g_{n\!}\!\in\!\mathcal{G}_n\}^N_{n=1}$. In \textit{Ad-hoCoop}, agents are even expected to learn more robust policies $\bm{\pi}$ that are adaptive to the team size $N$.

\subsection{$_{\!}$Memory-Augmented$_{\!}$ Communication$_{\!}$ for$_{\!}$ MAVN$_{\!\!\!\!}$}
\noindent \textbf{Overview.}  Our agents are connected by a low-bandwidth communication network without any central controller, through which they can coordinate their actions and behave as a group.  As shown in Fig.~\ref{fig:framework}(a), each agent mainly has two components: i) \textit{Map Building Module} (\S\ref{sec:MB})  and ii) \textit{Memory-Augmented Communication Module} (\S\ref{section:com_protocol}). For i), map building is a crucial step in navigation; an environment representation $\hat{\mathcal{S}}_n$ is estimated for path planning. For ii), each agent $n$ learns to generate useful information for communication and gather messages  $w_{n\!}\!\in\!\mathcal{W}_{n}$ from other agents simultaneously. It is equipped with a private memory for accurately storing and recalling past communication information. For each agent $n$, based on its private observation, estimated local map, communication information and target goal, its policy becomes $\bm{\pi}_{\theta_n\!}\!:\!\mathcal{O}_{n\!}\!\times\!\hat{\mathcal{S}}_{n\!}\!\times\!\mathcal{W}_{n\!}\!\times\!\mathcal{G}_{n\!}\mapsto\!\mathcal{A}_n$. See \S\ref{sec:ts} for the detailed definitions of $\mathcal{G}$, $\mathcal{A}$, and $\mathcal{O}$. Our whole system is built as trainable end-to-end; neural networks are used as approximations for policies and modules. Corresponding descriptions are presented below.

\subsubsection{Map Building Module}\label{sec:MB}
Learning environment layouts is crucial for  autonomous navigation~\cite{parisotto2017neural,zhang2017neural}.
Inspired by~\cite{chaplot2020object,chaplot2020learning}, we equip each agent with a map building module $\mathcal{F}^{\text{map}}$, which online estimates environment structures from agent's past percepts. For each agent $n$, we denote its pose as $p_n\!=\!(x_n, y_n, o_n)$, where ($x_n, y_n$) indicates its \textit{xy} co-ordinate  and $o_n$ represents its orientation in radians~\cite{chaplot2020object}. We set $p^0_n\!=\!(0,0,0)$ and let each agent build map on its ego-centric coordinate system.

For each agent $n$ at time step $t$, $\mathcal{F}^{\text{map}}$ takes its local observation $O^t_n$, current and last pose $p_n^{t-1:t}$, prior predicted map $S^{t-1}_n$ as input, and outputs an updated map:
\begin{equation}\small\label{eq:NCE}
S^{t}_n= \mathcal{F}^{\text{map}}(O^t_n, p_n^{t-1:t}, S^{t-1}_n).
\end{equation}
$S_n$ is a $(256\!+\!4)\!\times\!L\!\times\!L$ matrix where $L\!\times\!L$ denotes the map size. The first 256 channels store visual features over all the explored locations. To save the map size, the grid size is set as $9~\text{m}^{2} (3~\text{m}\!\times\!3~\text{m})$ in the physical world. The last four channels  include a probability map of obstacle and three binary maps storing explored area, agent past trajectory and current location, with $100~\text{cm}^{2} (10~\text{cm}\!\times\!10~\text{cm})$ gride size.
At the beginning of an episode, $S^0_n$ is initialized with all zeros and the agent $n$ is put in the center of $S^0_n$.

\subsubsection{Memory-Augmented Communication Module} \label{section:com_protocol}

In current communication based MARL algorithms~\cite{jiang2018learning, foerster2016learning, kim2019learning}, all the information generated in $t^{th}$-round communication are directly abandoned at the beginning of time step $t\!+\!1$ and new messages are generated for $(t\!+\!1)^{th}$-round communication. However, some discarded information may be useful in the future. In addition, the communication is mainly about agents' current observations, failing to involving their past experience explicitly. It is difficult if an agent wants to know another agent's status in a previous time step. We instead propose a memory-augmented communication strategy, that allows agent to accurately store and recall their past communication information. It is instantiated as a handshake communication based MARL~\cite{das2019tarmac,liu2020who2com}

\noindent \textbf{Handshake Strategy.} Three stages are involved in vanilla handshake communication, \ie,
\texttt{request}, \texttt{match}, and \texttt{select}, and each agent acts both \textit{requester} and \textit{supporter}:

\begin{itemize}[leftmargin=*]
	\setlength{\itemsep}{0pt}
	\setlength{\parsep}{-2pt}
	\setlength{\parskip}{-0pt}
	\setlength{\leftmargin}{-10pt}
	\vspace{-6pt}
	\item \texttt{request}.$_{\!}$ At$_{\!}$ the$_{\!}$ beginning$_{\!}$ of$_{\!}$ each$_{\!}$ time/communication$_{\!}$ step $t$, each agent $n$ first compresses its local observation $O^t_n$ to a \textit{query} $\bm{\mu}^t_n$, \textit{key} $\bm{\kappa}^t_n$, and \textit{value} $\bm{\upsilon}^t_n$:
        \begin{equation}\small
        \!\!\!\!\bm{\mu}^t_n \!=\! \mathcal{F}^{\text{query}}(O^t_n), ~~\bm{\kappa}^t_n  \!=\! \mathcal{F}^{\text{key}}(O^t_n), ~~\bm{\upsilon}^t_n  \!=\! \mathcal{F}^{\text{value}}(O^t_n),\!\!
        \end{equation}
        where $\bm{\mu}^t_{n\!}$ and $\bm{\kappa}^t_{n\!}$ are compact vectors while $\bm{\upsilon}^t_{n\!}$ is a high-dimension vector to fully preserve the information of $O^t_n$. Then the agent $n$ broadcasts $\bm{\mu}^t_{n\!}$ to other agents. As $\bm{\mu}^t_{n\!}$ is small-size, this only causes little bandwidth transmission.
\item \texttt{match}. Each of other agents derives a matching score $s_{n,m}$ between the received query $\bm{\mu}^t_n$ and its own key $\bm{\kappa}^t_m$:
        \begin{equation}\small
        s^t_{n,m} = \bm{\psi}(\bm{\mu}^t_n, \bm{\kappa}^t_m),~~~ \forall n\neq m,
        \end{equation}
        where $\bm{\psi}$ is a learnable similarity function. $s_{n,m}$ can be intuitively viewed as the importance of the information provided by the supporter $m$ for the requester $n$.
\item \texttt{select}. Some connections can be removed if their matching scores are small.  Then the requester $n$ collects information from its temporally connected supporters $m^t$ at time step $t$:
        \begin{equation}\small
        {\bm{\omega}}^t_n = \sum\nolimits_{m^t} s^t_{n,m^t}\bm{\upsilon}^t_{m^t},~~~ \forall n\neq m^t,
        \end{equation}
        where $\bm{\upsilon}^t_{m^t}$ is the value vector returned from the supporter $m^t$, and the integrated message ${\bm{\omega}}^t_n$ is used to help the requester $n$ take an action decision $a_n^t$.
\vspace{-3pt}
\end{itemize}

\noindent \textbf{Memory-Augmented Communication.} As shown in Fig.~\ref{fig:framework}(b), each agent $n$ maintains a private, external memory $\mathcal{M}^t_n$, which stores all its past generated communication information: $\{\bm{\upsilon}^{1:t-1}_{n}, \bm{\kappa}^{1:t-1}_n\}$. Then our memory-augmented handshake communication has the following four stages:
\begin{itemize}[leftmargin=*]
	\setlength{\itemsep}{0pt}
	\setlength{\parsep}{-2pt}
	\setlength{\parskip}{-0pt}
	\setlength{\leftmargin}{-10pt}
	\vspace{-6pt}
	\item \texttt{request}. At the beginning of each time
step $t$,  each agent $n$ generates \textit{query}, \textit{key}, and \textit{value} vectors:
        \vspace{-1pt}
        \begin{equation}\small
        \begin{aligned}
        \bm{\mu}^t_n \!=\! \mathcal{F}^{\text{query}}\big(O^t_n, S^t_{n}, \texttt{P}(\bm{\upsilon}^{1:t-1}_{n})\big),~~~~ ~~~~  \\
        \bm{\kappa}^t_n  \!=\! \mathcal{F}^{\text{key}}(O^t_n, S^t_{n}),~~~~
        \bm{\upsilon}^t_n \! = \!\mathcal{F}^{\text{value}}(O^t_n, S^t_{n}),
        \vspace{-1pt}
        \end{aligned}
        \end{equation}
        where \texttt{P}$(\cdot)$ is avg-pooling and both private observation $O^t_n$ and map $S^t_{n}$ are considered. In addition, past generated communication information, stored in $\mathcal{M}^t_n$, are also involved for generating a more reasonable $\bm{\mu}^t_n$.
\item \texttt{match}. Then the agent $n$ broadcasts $\bm{\mu}^t_n$ to other agents. Each of other agents looks up its current key $\bm{\kappa}^t_m$, and all the past self-generated keys $\bm{\kappa}^{1:t-1}_m$ stored in $\mathcal{M}^t_m$:
        \begin{equation}\small
        s^t_{n,m,1:t} = \bm{\psi}(\bm{\mu}^t_n, \bm{\kappa}^{1:t}_m),~~~ \forall n\neq m,
        \end{equation}
    Then the final matching score and returned message are:
        \begin{equation}\small\label{eq:xx}
        \!\!s^t_{n,m}\! =\! \frac{1}{t}\!\sum^t\nolimits_{t'=1\!}s^t_{n,m,t'},~~ \hat{\bm{\upsilon}}^t_m \!=\! \frac{1}{t}\!\sum^t\nolimits_{t'=1\!}s^t_{n,m,t'}\bm{\upsilon}^{t'}_m,\!\!
        \end{equation}
The agent $n$ also computes the correlations between the query $\bm{\mu}^t_n$ and all its current and past keys $\bm{\kappa}^{1:t}_n$ stored in $\mathcal{M}^t_n$:
        \begin{equation}\small
        s^t_{n,n,1:t} = \bm{\psi}(\bm{\mu}^t_n, \bm{\kappa}^{1:t}_n),
        \end{equation}
     Similar to Eq.~\ref{eq:xx}, we have:
        \begin{equation}\small
        \!\!s^t_{n,n}\! =\! \frac{1}{t}\!\sum^t\nolimits_{t'=1\!}s^t_{n,n,t'},~~ \hat{\bm{\upsilon}}^t_n \!=\! \frac{1}{t}\!\sum^t\nolimits_{t'=1\!}s^t_{n,n,t'}\bm{\upsilon}^{t'}_n.\!\!
        \end{equation}
\item \texttt{select}. To filter out some less connections, we apply an activation function $\Gamma(s^t_{n,m}, \eta)$ for each score $s^t_{n,m}$:
        \begin{equation}\small
        \Gamma(s^t_{n,m}, \eta) = \left\{
        \begin{aligned}
            &0,~~~~~~~~~~~~~~~~~~~~~~~~~\text{if~~} s^t_{n,m}\leq\eta\!\\
            &s^t_{n,m},~~~~~~~~~~~~~~~~~~~~\text{otherwise~~}\!\\
        \end{aligned}
    \right.\!\!
        \end{equation}
  As in~\cite{liu2020when2com}, only the supporters with non-zero $\Gamma$ scores are allowed to send messages to the requestor $n$ (\ie, learning \textit{who} to communicate). Moreover, when $\Gamma(s^t_{n,n}, \eta)\!=\!0$, that means the requestor $n$ thinks there is no need to establish any communication at this time step $t$, (\ie, learning \textit{when} to communicate). We set $\eta\!=\!1/N$ following~\cite{liu2020when2com}. If $\Gamma(s^t_{n,n}, \eta)\!>\!0$, the requester $n$ will  collect information from itself and other connected supporters $m_t$ (with $\Gamma(s^t_{n,m^t}, \eta)\!>\!0$):
        \begin{equation}\small
        {\bm{\omega}}^t_n = s^t_{n,n}\hat{\bm{\upsilon}}^t_n+\sum\nolimits_{m^t\neq n} s^t_{n,m^t}\hat{\bm{\upsilon}}^t_{m^t}.
        \end{equation}
  \item \texttt{store}. Next, the agent $n$ will store its value and key information $\{\bm{\upsilon}^{t}_{n}, \bm{\kappa}^{t}_n\}$ in the memory, getting $\mathcal{M}^t_{n}$ for the next round communication. In our current implementation, we simply assume the private memory for each agent is large enough to store all its generated communication information over the whole navigation episode. There are several ways to improve the efficiency. One can perform \texttt{store} operation in a certain interval of time, and/or build the memory as a stack with fixed capacity, and/or even learn a specific \texttt{write} operation to see if it is necessary to store current communication information by comparing its uniqueness with existing memory slots. We put this as a part of our future work.
\vspace{-3pt}
\end{itemize}

After $t^{th}$-round communication, each agent $n$ takes an navigation action through:
        \begin{equation}\small
        a^t_n\sim\bm{\pi}_{\theta_n\!}({O}^t_{n}, {S}^t_{n}, {\bm{\omega}}^t_n, {g}_{n}).
        \end{equation}

\subsection{Fully Decentralized Learning}\label{sec:PM}
At each time step $t$, each agent $n$ executes an individual action $a^t_{n\!}\!\sim\!\bm{\pi}_{\theta_n\!}$, with the joint goal of maximizing the average rewards of all the $N$ agents, \ie, $\frac{1}{N}\!\sum^N_{n=1\!}R_n$. We adopt an actor-critic model~\cite{mnih2016asynchronous} for training and consider it under a fully decentralized MARL setting~\cite{zhang2019multi,suttle2019multi,zhang2018fully} in which agents are connected via a time-varying communication network. Standard actor-critic methods consist of two models, an actor proposes an action and a critic gives a feedback/evaluation accordingly.  In PPO~\cite{schulman2017proximal}, neural networks are used as approximations of both the actor, represented by a policy function $\bm{\pi}_{\theta_n}$, and its corresponding critic, represented by a value function $V_{\phi_n\!}\!:\!\mathcal{O}_n\!\mapsto\!\mathbb{R}$. In order maximize $\mathbb{E}(R_n)$, we have the following policy loss (without clip operation):
        \begin{equation}\small
        \begin{aligned}
    \mathcal{L}(\theta_n)& = \mathbb{E}_{O_{n}^{t}, a_{n}^{t}}\left[\frac{\bm{\pi}_{\theta_n}(a_{n}^{t}|O_{n}^{t})}{\pi_{\theta_n^{old}}(a_{n}^{t}|O_{n}^{t})}A_{n}^{t}
    \right],\\
    \mathcal{L}(\phi_n) &= \mathbb{E}_{O_{n}^{t}} \left[V_{\phi_n}(O_{n}^{t}) - \hat{V}_{n}^{t} \right],
        \end{aligned}
        \end{equation}
where the advantage $A^t_n$ is estimated from $V_{\phi_n\!}$ and $R_n$~\cite{schulman2015high}, and $\hat{V}_{n}^{t}$ indicates the expected discounted return~\cite{williams1992simple}.

In this formulation, as there is no interaction between agents, the policies are learned independently. Previous MARL works~\cite{lowe2017multi,foerster2018counterfactual,yu2021surprising,iqbal2019actor,pesce2020improving} mainly address this by training multiple interacting agents in a \textit{Centralized Train and Decentralized Execution} manner. They learn decentralised policies with a centralized critic $V_{\phi\!}\!:\!\mathcal{O}_1\!\times\!\mathcal{O}_2\!\times\!\cdots\!\times\!\mathcal{O}_N\!\mapsto\!\mathbb{R}$
which collects all the observations to estimate joint state of all the agents. However, our communication based MARL method is essentially under a \textit{fully decentralized} setting~\cite{zhang2019multi,suttle2019multi,zhang2018fully}, in which agents are connected via a time-varying communication network. As agents can exchange information through communication, we learn an individual value function for each agent, dependent on both local observation and received message:
        \begin{equation}\small
        \begin{aligned}
    \mathcal{L}(\theta_n)& = \mathbb{E}_{O_{n}^{t}, a_{n}^{t}, {\bm{\omega}}^t_n}\left[\frac{\bm{\pi}_{\theta_n}(a_{n}^{t}|O_{n}^{t},{\bm{\omega}}^t_n)}{\pi_{\theta_n^{old}}(a_{n}^{t}|O_{n}^{t},{\bm{\omega}}^t_n)}A_{n}^{t}
    \right],\\
    \mathcal{L}(\phi_n) &= \mathbb{E}_{O_{n}^{t},{\bm{\omega}}^t_n} \left[V_{\phi_n}(O_{n}^{t},{\bm{\omega}}^t_n) - \hat{V}_{n}^{t} \right].
        \vspace{-3pt}
        \end{aligned}
        \end{equation}

%
%
%



\begin{table*}
    \centering
    \small
    \resizebox{1.\textwidth}{!}{
        \setlength\tabcolsep{6pt}
		\renewcommand\arraystretch{1.0}
        \begin{tabular}{c|l||cccc|cccc|cccc|cccc}
            \hline\thickhline
            \rowcolor{mygray}
 & &\multicolumn{4}{c|}{Easy (1.5-3 m)} & \multicolumn{4}{c|}{Medium (3-5 m)} & \multicolumn{4}{c|}{Hard (5-10 m)} & \multicolumn{4}{c}{Overall} \\
 \cline{3-18}
\cline{3-18}
\cline{3-18}
\rowcolor{mygray}
\multirow{-2}{*}{$N$} &\multirow{-2}{*}{Method} & SR$\uparrow$ & DTS$\downarrow$ & SSR$\uparrow$ & SPL$\uparrow$ & SR$\uparrow$ & DTS$\downarrow$ & SSR$\uparrow$ & SPL$\uparrow$ & SR$\uparrow$ & DTS$\downarrow$ & SSR$\uparrow$ & SPL$\uparrow$ & SR$\uparrow$ & DTS$\downarrow$ & SSR$\uparrow$ & SPL$\uparrow$\\
\hline\hline
\multirow{2}{*}{1} &Random             & 8.31  & 1.26 & 0.10 & 0.02 & 0.90 & 3.01 & 0.01 & 0.00 & 0.00 & 6.49 & 0.00 & 0.00 & 3.07 & 3.59 & 0.04 & 0.01 \\
&IL  & 37.10  & 1.22 & 2.64 & 0.13 & 10.10 & 2.75 & 0.25 & 0.05 & 1.80 & 5.81 & 0.07 & 0.01 & 16.33 & 3.26 & 0.98 & 0.06 \\
\hline\hline
\multirow{4}{*}{2}&IL    & 37.30  & 1.22 & 2.66 & 0.13 & 10.00 & 2.76 & 0.24 & 0.05 & 1.81 & 5.80 & 0.07 & 0.01 & 16.37 & 3.26 & 0.99 & 0.06  \\
&MARL \textit{w/o} com  & 39.10  & 1.19 & 2.98 & 0.19 & 12.10 & 2.61 & 0.53 & 0.06 & 1.89 & 5.76 & 0.07 & 0.01 & 17.69 & 3.18 & 1.11 & 0.08 \\
&MARL \textit{w/o} mem & 40.21  & 1.15 & 3.13 & 0.19 & 12.71 & 2.56 & 0.32 & 0.06 & 1.98 & 5.71 & 0.07 & 0.01 & 18.29 & 3.14 & 1.17 & 0.09 \\
&\textbf{Ours}  & \textbf{45.71}  & \textbf{1.05} & \textbf{3.53} & \textbf{0.22} & \textbf{16.55} & \textbf{2.42} & \textbf{0.41} & \textbf{0.08} & \textbf{2.31} & \textbf{5.54} & \textbf{0.09} & \textbf{0.01} &  \textbf{21.52} & \textbf{3.00} & \textbf{1.34} & \textbf{0.10} \\\hline\hline

\multirow{3}{*}{3} &MARL \textit{w/o} com  & 39.00  & 1.19 & 2.97 & 0.19 & 12.31 & 2.58 & 0.31 & 0.06 & 1.87 & 5.75 & 0.09 & 0.01 & 17.72 & 3.17 & 1.11 & 0.08 \\
&MARL \textit{w/o} mem & 40.81  & 1.14 & 3.21 & 0.20 & 12.92 & 2.54 & 0.34 & 0.06 & 2.01 & 5.71 & 0.07 & 0.01 & 18.57 & 3.13 & 1.20 & 0.08 \\
&\textbf{Ours} & \textbf{47.03} & \textbf{1.03} & \textbf{3.69} & \textbf{0.23} & \textbf{17.50} & \textbf{2.36} & \textbf{0.45} & \textbf{0.08} & \textbf{2.65} & \textbf{5.45} & \textbf{0.10} & \textbf{0.01} & \textbf{22.39} & \textbf{2.94} & \textbf{1.41} & \textbf{0.11} \\\hline\hline

\multirow{3}{*}{4}&MARL \textit{w/o} com  & 39.40  & 1.19 & 2.96 & 0.19 & 11.91 & 2.60 & 0.32 & 0.06 & 1.87 & 5.73 & 0.07 & 0.01 & 17.72 & 3.17 & 1.12 & 0.09 \\
&MARL \textit{w/o} mem & 41.02  & 1.14 & 3.23 & 0.20 & 13.10 & 2.52 & 0.34 & 0.06 & 2.01 & 5.71 & 0.07 & 0.01 & 18.70 & 3.12 & 1.21 & 0.09 \\
&\textbf{Ours}  & \textbf{48.01} & \textbf{1.01} & \textbf{3.75} & \textbf{0.24} & \textbf{18.40} & \textbf{2.31} & \textbf{0.48} & \textbf{0.09} & \textbf{2.91} & \textbf{5.40} & \textbf{0.11} & \textbf{0.01} & \textbf{23.10} & \textbf{2.90} & \textbf{1.45} & \textbf{0.11}\\
\hline
\end{tabular}
}
\captionsetup{font=small}
	\caption{\small Performance on \textit{CommonGoal} task with different team sizes ($N\!=\!2,3,4$) over our CommonGoal-CollaVN \texttt{test}. }
	\label{tab:commongoal}
\end{table*}

\begin{table*}
    \centering
    \small
    \resizebox{1.\textwidth}{!}{
        \setlength\tabcolsep{6pt}
		\renewcommand\arraystretch{1.0}
        \begin{tabular}{c|l||cccc|cccc|cccc|cccc}
            \hline\thickhline
            \rowcolor{mygray}
 & &\multicolumn{4}{c|}{Easy (1.5-3 m)} & \multicolumn{4}{c|}{Medium (3-5 m)} & \multicolumn{4}{c|}{Hard (5-10 m)} & \multicolumn{4}{c}{Overall} \\
 \cline{3-18}
\cline{3-18}
\cline{3-18}
\rowcolor{mygray}
\multirow{-2}{*}{$N$} &\multirow{-2}{*}{Method} & SR$\uparrow$ & DTS$\downarrow$ & SSR$\uparrow$ & SPL$\uparrow$ & SR$\uparrow$ & DTS$\downarrow$ & SSR$\uparrow$ & SPL$\uparrow$ & SR$\uparrow$ & DTS$\downarrow$ & SSR$\uparrow$ & SPL$\uparrow$ & SR$\uparrow$ & DTS$\downarrow$ & SSR$\uparrow$ & SPL$\uparrow$\\
\hline\hline
\multirow{3}{*}{2}&MARL \textit{w/o} com   & 39.02  & 1.19 & 2.99 & 0.19  & 12.00 & 2.59 & 0.30 & 0.06  & 1.88 & 5.75 & 0.07 & 0.01  & 17.63 & 3.18 & 1.12 & 0.09  \\
&MARL \textit{w/o} mem  & 39.86  & 1.17 & 3.05 & 0.19  & 12.40 & 2.58 & 0.31 & 0.06  & 1.94 & 5.72 & 0.07 & 0.01  & 18.06 & 3.16 & 1.14 & 0.09  \\
&\textbf{Ours}  & \textbf{44.50}  & \textbf{1.07} & \textbf{3.49} & \textbf{0.21}  & \textbf{15.95} & \textbf{2.45} & \textbf{0.40} & \textbf{0.08}  & \textbf{2.26} & \textbf{5.58} & \textbf{0.08} & \textbf{0.01}  & \textbf{20.90} & \textbf{3.03} & \textbf{1.32} & \textbf{0.10}  \\\hline\hline
\multirow{3}{*}{3}&MARL \textit{w/o} com & 39.08  & 1.19 & 2.98 & 0.19  & 12.12 & 2.58 & 0.30 & 0.06  & 1.87 & 5.74 & 0.07 & 0.01  & 17.69 & 3.17 & 1.12 & 0.09 \\
&MARL \textit{w/o} mem & 40.20  & 1.16 & 3.08 & 0.19  & 12.65 & 2.56 & 0.32 & 0.06  & 1.96 & 5.71 & 0.07 & 0.01  & 18.27 & 3.14 & 1.16 & 0.09 \\
&\textbf{Ours}  & \textbf{45.83}  & \textbf{1.05} & \textbf{3.61} & \textbf{0.22}  & \textbf{17.00} & \textbf{2.38} & \textbf{0.43} & \textbf{0.08}  & \textbf{2.57} & \textbf{5.49} & \textbf{0.09} & \textbf{0.01}  & \textbf{21.80} & \textbf{2.97} & \textbf{1.38} & \textbf{0.10} \\\hline\hline
\multirow{3}{*}{4}&MARL \textit{w/o} com & 39.14  & 1.19 & 2.95 & 0.19  & 12.01 & 2.59 & 0.31 & 0.06  & 1.88 & 5.73 & 0.07 & 0.01  & 17.68 & 3.17 & 1.11 & 0.09\\
&MARL \textit{w/o} mem & 40.60  & 1.15 & 3.13 & 0.20  & 12.85 & 2.54 & 0.33 & 0.06  & 1.98 & 5.71 & 0.07 & 0.01  & 18.47 & 3.13 & 1.18 & 0.09 \\
&\textbf{Ours}
& \textbf{46.79}  & \textbf{1.03} & \textbf{3.71} & \textbf{0.23}  & \textbf{17.82} & \textbf{2.33} & \textbf{0.46} & \textbf{0.09}  & \textbf{2.80} & \textbf{5.42} & \textbf{0.10} & \textbf{0.01}  & \textbf{22.47} & \textbf{2.93} & \textbf{1.42} & \textbf{0.11}\\
\hline
\end{tabular}
}
\captionsetup{font=small}
	\caption{\small Performance on \textit{SpecificGoal} task with different team sizes ($N\!=\!2,3,4$) over our Specific-CollaVN \texttt{test}.}
\label{tab:specific goal}
\end{table*}

\begin{table*}[!tbh]
    \centering
    \small
    \resizebox{1.\textwidth}{!}{
        \setlength\tabcolsep{6pt}
		\renewcommand\arraystretch{1.0}
        \begin{tabular}{c|l||cccc|cccc|cccc|cccc}
            \hline\thickhline
            \rowcolor{mygray}
 & &\multicolumn{4}{c|}{Easy (1.5-3 m)} & \multicolumn{4}{c|}{Medium (3-5 m)} & \multicolumn{4}{c|}{Hard (5-10 m)} & \multicolumn{4}{c}{Overall} \\
 \cline{3-18}
\cline{3-18}
\cline{3-18}
\rowcolor{mygray}
\multirow{-2}{*}{$N$} &\multirow{-2}{*}{Method} & SR$\uparrow$ & DTS$\downarrow$ & SSR$\uparrow$ & SPL$\uparrow$ & SR$\uparrow$ & DTS$\downarrow$ & SSR$\uparrow$ & SPL$\uparrow$ & SR$\uparrow$ & DTS$\downarrow$ & SSR$\uparrow$ & SPL$\uparrow$ & SR$\uparrow$ & DTS$\downarrow$ & SSR$\uparrow$ & SPL$\uparrow$\\
\hline\hline
\multirow{3}{*}{2$\rightarrow$3}
&MARL \textit{w/o} com    & 39.10  & 1.19 & 2.97 & 0.19 & 12.3 & 2.58 & 0.32 & 0.06  & 1.88 & 5.75 & 0.07 & 0.01  & 17.76 & 3.17 & 1.12 & 0.09  \\
&MARL \textit{w/o} mem  & 40.50  & 1.15 & 3.17 & 0.20 & 12.87 & 2.55 & 0.34 & 0.06 & 1.99 & 5.72 & 0.07 & 0.01 & 18.45 & 3.14 & 1.19 & 0.09 \\
&\textbf{Ours}    & \textbf{46.66}  & \textbf{1.03} & \textbf{3.68} & \textbf{0.22}  & \textbf{17.30} & \textbf{2.37} & \textbf{0.45} & \textbf{0.08}  & \textbf{2.60} & \textbf{5.47} & \textbf{0.10} & \textbf{0.01}  & \textbf{22.17} & \textbf{2.96} & \textbf{1.41} & \textbf{0.11}  \\\hline\hline
\multirow{3}{*}{3$\rightarrow$2}
&MARL \textit{w/o} com   & 39.00  & 1.19 & 2.98 & 0.19  & 12.20 & 2.60 & 0.30 & 0.06  & 1.89 & 5.76 & 0.07 & 0.01  & 17.70 & 3.18 & 1.12 & 0.09  \\
&MARL \textit{w/o} mem & 39.90  & 1.17 & 3.10 & 0.19 & 12.40 & 2.58 & 0.31 & 0.06 & 1.96 & 5.71 & 0.07 & 0.01 & 18.08 & 3.15 & 1.16 & 0.09 \\
&\textbf{Ours} & \textbf{45.20}  & \textbf{1.06} & \textbf{3.51} & \textbf{0.22}  & \textbf{16.10} & \textbf{2.43} & \textbf{0.41} & \textbf{0.08}  & \textbf{2.27} & \textbf{5.58} & \textbf{0.08} & \textbf{0.01}  & \textbf{21.19} & \textbf{3.02}  & \textbf{1.33} & \textbf{0.10}  \\
\hline
\end{tabular}
}
\captionsetup{font=small}
	\caption{\small Performance on \textit{Ad-hoCoop} task with two ad-hoc team-play settings ($N\!:\!2\!\rightarrow\!3$ and $N\!:\!3\!\rightarrow\!2$) over our Ad-Hoc-CollaVN \texttt{test}.}
\label{tab:ad_hoc}
\end{table*}

\section{Experiments}
We conduct experiments on the three MAVN tasks, \ie, \textit{CommonGoal}, \textit{SpecificGoal}, \textit{Ad-hoCoop} on the corresponding subdatasets of CollaVN, \ie, CommonGoal-CollaVN, SpecificGoal-CollaVN,  and Ad-Hoc-CollaVN.
\subsection{Experimental Setup}

\noindent\textbf{Network Details.}  The embedding of local visual observation $O_n$ is from a pretrained ResNet18~\cite{he2016deep} (compressed in to 1024-\textit{d}). The implementation of map building module $\mathcal{F}^{\text{map}}$ (\S\ref{sec:MB}) mainly follows~\cite{chaplot2020object,chaplot2020learning} and the map size is $38.4 m\!\times\!38.4 m$ in the physical world.  Query $\bm{\mu}$, key $\bm{\kappa}$, and value $\bm{\upsilon}$ in \S\ref{section:com_protocol} are 256-\textit{d}, 256-\textit{d}, 2048-\textit{d} vectors, respectively. All the functions in \S\ref{section:com_protocol} are one-layer MLPs. The policy $\bm{\pi}$ and value function $V$ are GRUs~\cite{chung2014empirical}.

\noindent\textbf{Evaluation.}  For different MALN tasks, we adopt \texttt{test} sets of corresponding sub-datasets in CollaVN for benchmarking, measured by (\S\ref{sec:3.3})
SR,  DTS, SPL and SSR. The maximum episode length is set as $T\!=\!80$ steps.

\noindent\textbf{Training.} Our model was trained using dense reward of geodesic distance reduced to the goal location and constant -0.05 step penalty to collision.

\noindent\textbf{Reproducibility.} Our model is implemented in PyTorch and trained on four NVIDIA Tesla V100 GPUs with a 32GB memory per-card.  To reveal full details of our algorithm, our implementations are released.

\subsection{Baseline}\label{sec:5.1}
Following baselines are used in our experiments (their sub-modules are the same with ours unless specified):

\begin{itemize}[leftmargin=*]
	\setlength{\itemsep}{0pt}
	\setlength{\parsep}{-2pt}
	\setlength{\parskip}{-0pt}
	\setlength{\leftmargin}{-10pt}
	\vspace{-6pt}
	\item \textbf{Random walk} is the simplest heuristic for navigation: agent randomly selects an action at each step.

	\item \textbf{IL} learns an off policy by imitation learning (IL). One agent is trained and several copies are used for testing.


	\item \textbf{MARL \textit{w/o} com} learns target-driven MARL policy without explicit communication, using IPPO~\cite{schroeder2020independent}.

	\item\textbf{MARL \textit{w/o} mem} can be viewed as a variant of our model without using memory.
\vspace{-3pt}
\end{itemize}

\subsection{Experimental Results} \label{section:result}
\noindent\textbf{Performance on \textit{CommonGoal} Task.} For the settings with different agent numbers, \ie, $N\!=\!2,3,4$ on the CommonGoal-CollaVN sub-dataset, all the benchmarked models are trained on the corresponding \texttt{train} sets and tested on the corresponding \texttt{test} sets. The results are summarized in Table~\ref{tab:commongoal}. As seen, our model outperforms over all other baselines across all the metrics. Interestingly, we can also observe IL based baseline, \ie, IL-M, even shows comparable performance with a MARL baseline, \ie, MARL \textit{w/o} com, revealing the difficulty of this task.

\noindent\textbf{Performance on \textit{SpecificGoal} Task.} Also, for different agent numbers on the SpecificGoal-CollaVN sub-dataset, we train and evaluate the models on the corresponding \texttt{train} and \texttt{test} sets. From Table~\ref{tab:specific goal} we can find that our method sill gets better performance. However, compared with the results in  Table~\ref{tab:commongoal}, all the methods suffer from performance drop, which indicates that it is harder to learn communication when agents perform different tasks.

\noindent\textbf{Performance on \textit{Ad-hoCoop} Task.} For $N\!\!:\!2\!\rightarrow\!3$ setting in the Ad-Hoc-CollaVN dataset, each of \texttt{train} scenes contains two agents while each of \texttt{test} scenes have three agents. Similarly, for $N\!:\!3\!\rightarrow\!2$ setting, each of \texttt{train} scenes contains three  agents while each of \texttt{test} scenes have two agents. For the two settings, all the methods are trained and test on the corresponding \texttt{train} and \texttt{test} sets. The results in Table~\ref{tab:ad_hoc} show again our model performs better on the two ad-hoc team play settings, mainly due to our fully decentralized learning strategy.

\noindent\textbf{Effect on Memory-Based Communication.} Compared with MARL \textit{w/o} mem, our model shows significantly better performance over all the three MAVN settings. MARL \textit{w/o} mem only shows limited improvements over MARL \textit{w/o} com. These observations suggest the importance of long-term memory in MAVN.

\section{Conclusion}
A new dataset was introduced for multi-agent collaborative navigation in complex environments. A memory-based communication model was presented to address the reuse of past communication information and facilitate cooperation. Our experiments showed that there is still huge room for further improvement. This work is expected to inspire more future efforts towards this promising direction.

{\small
\bibliographystyle{ieee_fullname}
\bibliography{egbib}

\begin{thebibliography}{10}\itemsep=-1pt

\bibitem{albrecht2018autonomous}
Stefano~V Albrecht and Peter Stone.
\newblock Autonomous agents modelling other agents: A comprehensive survey and
  open problems.
\newblock {\em Artificial Intelligence}, 258:66--95, 2018.

\bibitem{anderson2018evaluation}
Peter Anderson, Angel Chang, Devendra~Singh Chaplot, Alexey Dosovitskiy,
  Saurabh Gupta, Vladlen Koltun, Jana Kosecka, Jitendra Malik, Roozbeh
  Mottaghi, Manolis Savva, et~al.
\newblock On evaluation of embodied navigation agents.
\newblock {\em arXiv preprint arXiv:1807.06757}, 2018.

\bibitem{anderson2018vision}
Peter Anderson, Qi Wu, Damien Teney, Jake Bruce, Mark Johnson, Niko
  S{\"u}nderhauf, Ian Reid, Stephen Gould, and Anton van~den Hengel.
\newblock Vision-and-language navigation: Interpreting visually-grounded
  navigation instructions in real environments.
\newblock In {\em CVPR}, 2018.

\bibitem{armeni20163d}
Iro Armeni, Ozan Sener, Amir~R Zamir, Helen Jiang, Ioannis Brilakis, Martin
  Fischer, and Silvio Savarese.
\newblock 3d semantic parsing of large-scale indoor spaces.
\newblock In {\em CVPR}, 2016.

\bibitem{bansal2018emergent}
Trapit Bansal, Jakub Pachocki, Szymon Sidor, Ilya Sutskever, and Igor Mordatch.
\newblock Emergent complexity via multi-agent competition.
\newblock {\em arXiv preprint arXiv:1710.03748}, 2017.

\bibitem{barrett2011empirical}
Samuel Barrett, Peter Stone, and Sarit Kraus.
\newblock Empirical evaluation of ad hoc teamwork in the pursuit domain.
\newblock In {\em AAMAS}, pages 567--574, 2011.

\bibitem{openai2019dota}
Christopher Berner, Greg Brockman, Brooke Chan, Vicki Cheung, Shariq Hashme,
  Chris Hesse, Rafal J{\'o}zefowicz, Scott Gray, Catherine Olsson, Jakub
  Pachocki, et~al.
\newblock Dota 2 with large scale deep reinforcement learning.
\newblock {\em arXiv preprint arXiv:1912.06680}, 2019.

\bibitem{borenstein1991vector}
Johann Borenstein and Yoram Koren.
\newblock The vector field histogram-fast obstacle avoidance for mobile robots.
\newblock {\em IEEE Transactions on Robotics and Automation}, 7(3):278--288,
  1991.

\bibitem{bowling2005coordination}
Michael Bowling and Peter McCracken.
\newblock Coordination and adaptation in impromptu teams.
\newblock In {\em AAAI}, 2005.

\bibitem{brodeur2017home}
Simon Brodeur, Ethan Perez, Ankesh Anand, Florian Golemo, Luca Celotti, Florian
  Strub, Jean Rouat, Hugo Larochelle, and Aaron Courville.
\newblock Home: A household multimodal environment.
\newblock {\em arXiv preprint arXiv:1711.11017}, 2017.

\bibitem{busoniu2008comprehensive}
Lucian Busoniu, Robert Babuska, and Bart De~Schutter.
\newblock A comprehensive survey of multiagent reinforcement learning.
\newblock {\em IEEE Transactions on Systems, Man, and Cybernetics, Part C
  (Applications and Reviews)}, 38(2):156--172, 2008.

\bibitem{canaan2019diverse}
Rodrigo Canaan, Julian Togelius, Andy Nealen, and Stefan Menzel.
\newblock Diverse agents for ad-hoc cooperation in hanabi.
\newblock In {\em IEEE Conference on Games}, 2019.

\bibitem{chang2017matterport3d}
Angel Chang, Angela Dai, Thomas Funkhouser, Maciej Halber, Matthias Niebner,
  Manolis Savva, Shuran Song, Andy Zeng, and Yinda Zhang.
\newblock Matterport3d: Learning from rgb-d data in indoor environments.
\newblock In {\em 3DV}, 2018.

\bibitem{chaplot2020object}
Devendra~Singh Chaplot, Dhiraj Gandhi, Abhinav Gupta, and Ruslan Salakhutdinov.
\newblock Object goal navigation using goal-oriented semantic exploration.
\newblock {\em arXiv preprint arXiv:2007.00643}, 2020.

\bibitem{chaplot2020learning}
Devendra~Singh Chaplot, Dhiraj Gandhi, Saurabh Gupta, Abhinav Gupta, and Ruslan
  Salakhutdinov.
\newblock Learning to explore using active neural slam.
\newblock In {\em ICLR}, 2020.

\bibitem{chaplot2020neural}
Devendra~Singh Chaplot, Ruslan Salakhutdinov, Abhinav Gupta, and Saurabh Gupta.
\newblock Neural topological slam for visual navigation.
\newblock In {\em CVPR}, 2020.

\bibitem{chen2020soundspaces}
Changan Chen, Unnat Jain, Carl Schissler, Sebastia Vicenc~Amengual Gari, Ziad
  Al-Halah, Vamsi~Krishna Ithapu, Philip Robinson, and Kristen Grauman.
\newblock Soundspaces: Audio-visual navigation in 3d environments.
\newblock In {\em ECCV}, 2020.

\bibitem{chen2019touchdown}
Howard Chen, Alane Suhr, Dipendra Misra, Noah Snavely, and Yoav Artzi.
\newblock Touchdown: Natural language navigation and spatial reasoning in
  visual street environments.
\newblock In {\em CVPR}, 2019.

\bibitem{chung2014empirical}
Junyoung Chung, Caglar Gulcehre, KyungHyun Cho, and Yoshua Bengio.
\newblock Empirical evaluation of gated recurrent neural networks on sequence
  modeling.
\newblock {\em arXiv preprint arXiv:1412.3555}, 2014.

\bibitem{das2018embodied}
Abhishek Das, Samyak Datta, Georgia Gkioxari, Stefan Lee, Devi Parikh, and
  Dhruv Batra.
\newblock Embodied question answering.
\newblock In {\em CVPR}, 2018.

\bibitem{das2019tarmac}
Abhishek Das, Th{\'e}ophile Gervet, Joshua Romoff, Dhruv Batra, Devi Parikh,
  Mike Rabbat, and Joelle Pineau.
\newblock Tarmac: Targeted multi-agent communication.
\newblock In {\em ICML}, 2019.

\bibitem{deitke2020robothor}
Matt Deitke, Winson Han, Alvaro Herrasti, Aniruddha Kembhavi, Eric Kolve,
  Roozbeh Mottaghi, Jordi Salvador, Dustin Schwenk, Eli VanderBilt, Matthew
  Wallingford, et~al.
\newblock Robothor: An open simulation-to-real embodied ai platform.
\newblock In {\em CVPR}, 2020.

\bibitem{deng2020evolving}
Zhiwei Deng, Karthik Narasimhan, and Olga Russakovsky.
\newblock Evolving graphical planner: Contextual global planning for
  vision-and-language navigation.
\newblock {\em arXiv preprint arXiv:2007.05655}, 2020.

\bibitem{foerster2016learning}
Jakob Foerster, Ioannis~Alexandros Assael, Nando de Freitas, and Shimon
  Whiteson.
\newblock Learning to communicate with deep multi-agent reinforcement learning.
\newblock In D. Lee, M. Sugiyama, U. Luxburg, I. Guyon, and R. Garnett,
  editors, {\em NeurIPS}, 2016.

\bibitem{foerster2018counterfactual}
Jakob Foerster, Gregory Farquhar, Triantafyllos Afouras, Nantas Nardelli, and
  Shimon Whiteson.
\newblock Counterfactual multi-agent policy gradients.
\newblock In {\em AAAI}, 2018.

\bibitem{foerster2017stabilising}
Jakob Foerster, Nantas Nardelli, Gregory Farquhar, Triantafyllos Afouras,
  Philip~HS Torr, Pushmeet Kohli, and Shimon Whiteson.
\newblock Stabilising experience replay for deep multi-agent reinforcement
  learning.
\newblock In {\em ICML}, 2017.

\bibitem{fox2000probabilistic}
Dieter Fox, Wolfram Burgard, Hannes Kruppa, and Sebastian Thrun.
\newblock A probabilistic approach to collaborative multi-robot localization.
\newblock {\em Autonomous robots}, 8(3):325--344, 2000.

\bibitem{goertzel2007artificial}
Ben Goertzel and Cassio Pennachin.
\newblock {\em Artificial general intelligence}, volume~2.
\newblock Springer, 2007.

\bibitem{gupta2017cooperative}
Jayesh~K Gupta, Maxim Egorov, and Mykel Kochenderfer.
\newblock Cooperative multi-agent control using deep reinforcement learning.
\newblock In {\em International Conference on Autonomous Agents and Multiagent
  Systems}, 2017.

\bibitem{gupta2017cognitive}
Saurabh Gupta, James Davidson, Sergey Levine, Rahul Sukthankar, and Jitendra
  Malik.
\newblock Cognitive mapping and planning for visual navigation.
\newblock In {\em CVPR}, 2017.

\bibitem{hartley2003multiple}
Richard Hartley and Andrew Zisserman.
\newblock {\em Multiple view geometry in computer vision}.
\newblock Cambridge university press, 2003.

\bibitem{hausknecht2016half}
Matthew Hausknecht, Prannoy Mupparaju, Sandeep Subramanian, Shivaram
  Kalyanakrishnan, and Peter Stone.
\newblock Half field offense: An environment for multiagent learning and ad hoc
  teamwork.
\newblock In {\em AAMAS Adaptive Learning Agents (ALA) Workshop}, 2016.

\bibitem{he2016deep}
Kaiming He, Xiangyu Zhang, Shaoqing Ren, and Jian Sun.
\newblock Deep residual learning for image recognition.
\newblock In {\em CVPR}, 2016.

\bibitem{Hernandez_Leal_2019}
Pablo Hernandez-Leal, Bilal Kartal, and Matthew~E Taylor.
\newblock A survey and critique of multiagent deep reinforcement learning.
\newblock {\em Autonomous Agents and Multi-Agent Systems}, 33(6):750--797,
  2019.

\bibitem{hong2017deep}
Zhang-Wei Hong, Shih-Yang Su, Tzu-Yun Shann, Yi-Hsiang Chang, and Chun-Yi Lee.
\newblock A deep policy inference q-network for multi-agent systems.
\newblock {\em arXiv preprint arXiv:1712.07893}, 2017.

\bibitem{iqbal2019actor}
Shariq Iqbal and Fei Sha.
\newblock Actor-attention-critic for multi-agent reinforcement learning.
\newblock In {\em ICML}, 2019.

\bibitem{jaderberg2019human}
Max Jaderberg, Wojciech~M Czarnecki, Iain Dunning, Luke Marris, Guy Lever,
  Antonio~Garcia Castaneda, Charles Beattie, Neil~C Rabinowitz, Ari~S Morcos,
  Avraham Ruderman, et~al.
\newblock Human-level performance in 3d multiplayer games with population-based
  reinforcement learning.
\newblock {\em Science}, 364(6443):859--865, 2019.

\bibitem{jain2020cordial}
Unnat Jain, Luca Weihs, Eric Kolve, Ali Farhadi, Svetlana Lazebnik, Aniruddha
  Kembhavi, and Alexander Schwing.
\newblock A cordial sync: Going beyond marginal policies for multi-agent
  embodied tasks.
\newblock In {\em ECCV}, 2020.

\bibitem{jain2019two}
Unnat Jain, Luca Weihs, Eric Kolve, Mohammad Rastegari, Svetlana Lazebnik, Ali
  Farhadi, Alexander~G Schwing, and Aniruddha Kembhavi.
\newblock Two body problem: Collaborative visual task completion.
\newblock In {\em CVPR}, 2019.

\bibitem{Jiang2020Graph}
Jiechuan Jiang, Chen Dun, Tiejun Huang, and Zongqing Lu.
\newblock Graph convolutional reinforcement learning.
\newblock In {\em ICLR}, 2020.

\bibitem{jiang2018learning}
Jiechuan Jiang and Zongqing Lu.
\newblock Learning attentional communication for multi-agent cooperation.
\newblock In {\em NeurIPS}, 2018.

\bibitem{kantaros2015distributed}
Yiannis Kantaros, Michalis Thanou, and Anthony Tzes.
\newblock Distributed coverage control for concave areas by a heterogeneous
  robot--swarm with visibility sensing constraints.
\newblock {\em Automatica}, 53:195--207, 2015.

\bibitem{kim2019learning}
Daewoo Kim, Sangwoo Moon, David Hostallero, Wan~Ju Kang, Taeyoung Lee,
  Kyunghwan Son, and Yung Yi.
\newblock Learning to schedule communication in multi-agent reinforcement
  learning.
\newblock {\em arXiv preprint arXiv:1902.01554}, 2019.

\bibitem{kim1999symbolic}
Dongsung Kim and Ramakant Nevatia.
\newblock Symbolic navigation with a generic map.
\newblock {\em Autonomous Robots}, 6(1):69--88, 1999.

\bibitem{kolve2017ai2}
Eric Kolve, Roozbeh Mottaghi, Winson Han, Eli VanderBilt, Luca Weihs, Alvaro
  Herrasti, Daniel Gordon, Yuke Zhu, Abhinav Gupta, and Ali Farhadi.
\newblock Ai2-thor: An interactive 3d environment for visual ai.
\newblock {\em arXiv preprint arXiv:1712.05474}, 2017.

\bibitem{pytorchrl}
Ilya Kostrikov.
\newblock Pytorch implementations of reinforcement learning algorithms.
\newblock \url{https://github.com/ikostrikov/pytorch-a2c-ppo-acktr-gail}, 2018.

\bibitem{kuipers1978modeling}
Benjamin Kuipers.
\newblock Modeling spatial knowledge.
\newblock {\em Cognitive Science}, 2(2):129--153, 1978.

\bibitem{landemore2008democratic}
Helene~Emilie Landemore.
\newblock {\em Democratic reason: Politics, collective intelligence, and the
  rule of the many}.
\newblock PhD thesis, Harvard University, 2008.

\bibitem{lazaridou2017multiagent}
Angeliki Lazaridou, Alexander Peysakhovich, and Marco Baroni.
\newblock Multi-agent cooperation and the emergence of (natural) language.
\newblock In {\em ICLR}, 2017.

\bibitem{lee2018gated}
Lisa Lee, Emilio Parisotto, Devendra~Singh Chaplot, Eric Xing, and Ruslan
  Salakhutdinov.
\newblock Gated path planning networks.
\newblock In {\em ICML}, 2018.

\bibitem{littman1994markov}
Michael~L Littman.
\newblock Markov games as a framework for multi-agent reinforcement learning.
\newblock In {\em Machine Learning Proceedings}, pages 157--163. 1994.

\bibitem{liu2016multirobot}
Yugang Liu and Goldie Nejat.
\newblock Multirobot cooperative learning for semiautonomous control in urban
  search and rescue applications.
\newblock {\em Journal of Field Robotics}, 33(4):512--536, 2016.

\bibitem{liu2020when2com}
Yen-Cheng Liu, Junjiao Tian, Nathaniel Glaser, and Zsolt Kira.
\newblock When2com: multi-agent perception via communication graph grouping.
\newblock In {\em CVPR}, 2020.

\bibitem{liu2020who2com}
Yen-Cheng Liu, Junjiao Tian, Chih-Yao Ma, Nathan Glaser, Chia-Wen Kuo, and
  Zsolt Kira.
\newblock Who2com: Collaborative perception via learnable handshake
  communication.
\newblock In {\em ICRA}, 2020.

\bibitem{lowe2017multi}
Ryan Lowe, Yi Wu, Aviv Tamar, Jean Harb, Pieter Abbeel, and Igor Mordatch.
\newblock Multi-agent actor-critic for mixed cooperative-competitive
  environments.
\newblock {\em arXiv preprint arXiv:1706.02275}, 2017.

\bibitem{matignon2012independent}
Laetitia Matignon, Guillaume~J Laurent, and Nadine Le~Fort-Piat.
\newblock Independent reinforcement learners in cooperative markov games: a
  survey regarding coordination problems.
\newblock {\em Knowledge Engineering Review}, 27(1):1--31, 2012.

\bibitem{miah2014nonuniform}
Suruz Miah, Bao Nguyen, Fran{\c{c}}ois-Alex Bourque, and Davide Spinello.
\newblock Nonuniform deployment of autonomous agents in harbor-like
  environments.
\newblock {\em Unmanned Systems}, 2(04):377--389, 2014.

\bibitem{mnih2016asynchronous}
Volodymyr Mnih, Adria~Puigdomenech Badia, Mehdi Mirza, Alex Graves, Timothy
  Lillicrap, Tim Harley, David Silver, and Koray Kavukcuoglu.
\newblock Asynchronous methods for deep reinforcement learning.
\newblock In {\em ICML}, 2016.

\bibitem{mordatch2018emergence}
Igor Mordatch and Pieter Abbeel.
\newblock Emergence of grounded compositional language in multi-agent
  populations.
\newblock In {\em AAAI}, 2018.

\bibitem{pyrobot2019}
Adithyavairavan Murali, Tao Chen, Kalyan~Vasudev Alwala, Dhiraj Gandhi, Lerrel
  Pinto, Saurabh Gupta, and Abhinav Gupta.
\newblock Pyrobot: An open-source robotics framework for research and
  benchmarking.
\newblock {\em arXiv preprint arXiv:1906.08236}, 2019.

\bibitem{palmer2017lenient}
Gregory Palmer, Karl Tuyls, Daan Bloembergen, and Rahul Savani.
\newblock Lenient multi-agent deep reinforcement learning.
\newblock {\em arXiv preprint arXiv:1707.04402}, 2017.

\bibitem{panait2005cooperative}
Liviu Panait and Sean Luke.
\newblock Cooperative multi-agent learning: The state of the art.
\newblock {\em Autonomous Agents and Multi-Agent Systems}, 11(3):387--434,
  2005.

\bibitem{parisotto2017neural}
Emilio Parisotto and Ruslan Salakhutdinov.
\newblock Neural map: Structured memory for deep reinforcement learning.
\newblock In {\em ICLR}, 2019.

\bibitem{peng2017multiagent}
Peng Peng, Ying Wen, Yaodong Yang, Quan Yuan, Zhenkun Tang, Haitao Long, and
  Jun Wang.
\newblock Multiagent bidirectionally-coordinated nets: Emergence of human-level
  coordination in learning to play starcraft combat games.
\newblock {\em arXiv preprint arXiv:1703.10069}, 2017.

\bibitem{Peng_2018}
Xue~Bin Peng, Marcin Andrychowicz, Wojciech Zaremba, and Pieter Abbeel.
\newblock Sim-to-real transfer of robotic control with dynamics randomization.
\newblock In {\em ICRA}, 2018.

\bibitem{pesce2020improving}
Emanuele Pesce and Giovanni Montana.
\newblock Improving coordination in small-scale multi-agent deep reinforcement
  learning through memory-driven communication.
\newblock {\em Machine Learning}, pages 1--21, 2020.

\bibitem{raghu2018can}
Maithra Raghu, Alex Irpan, Jacob Andreas, Bobby Kleinberg, Quoc Le, and Jon
  Kleinberg.
\newblock Can deep reinforcement learning solve erdos-selfridge-spencer games?
\newblock In {\em ICML}, 2018.

\bibitem{raileanu2018modeling}
Roberta Raileanu, Emily Denton, Arthur Szlam, and Rob Fergus.
\newblock Modeling others using oneself in multi-agent reinforcement learning.
\newblock In {\em ICML}, 2018.

\bibitem{rashid2018qmix}
Tabish Rashid, Mikayel Samvelyan, Christian Schroeder, Gregory Farquhar, Jakob
  Foerster, and Shimon Whiteson.
\newblock Qmix: Monotonic value function factorisation for deep multi-agent
  reinforcement learning.
\newblock In {\em ICML}, 2018.

\bibitem{samvelyan2019starcraft}
Mikayel Samvelyan, Tabish Rashid, Christian~Schroeder De~Witt, Gregory
  Farquhar, Nantas Nardelli, Tim~GJ Rudner, Chia-Man Hung, Philip~HS Torr,
  Jakob Foerster, and Shimon Whiteson.
\newblock The starcraft multi-agent challenge.
\newblock {\em arXiv preprint arXiv:1902.04043}, 2019.

\bibitem{savva2019habitat}
Manolis Savva, Abhishek Kadian, Oleksandr Maksymets, Yili Zhao, Erik Wijmans,
  Bhavana Jain, Julian Straub, Jia Liu, Vladlen Koltun, Jitendra Malik, et~al.
\newblock Habitat: A platform for embodied ai research.
\newblock In {\em ICCV}, 2019.

\bibitem{schroeder2020independent}
Christian Schroeder~de Witt, Tarun Gupta, Denys Makoviichuk, Viktor
  Makoviychuk, Philip~HS Torr, Mingfei Sun, and Shimon Whiteson.
\newblock Is independent learning all you need in the starcraft multi-agent
  challenge?
\newblock {\em arXiv preprint arXiv:2011.09533}, 2020.

\bibitem{schulman2015high}
John Schulman, Philipp Moritz, Sergey Levine, Michael Jordan, and Pieter
  Abbeel.
\newblock High-dimensional continuous control using generalized advantage
  estimation.
\newblock {\em arXiv preprint arXiv:1506.02438}, 2015.

\bibitem{schulman2017proximal}
John Schulman, Filip Wolski, Prafulla Dhariwal, Alec Radford, and Oleg Klimov.
\newblock Proximal policy optimization algorithms.
\newblock {\em arXiv preprint arXiv:1707.06347}, 2017.

\bibitem{shen2020igibson}
Bokui Shen, Fei Xia, Chengshu Li, Roberto Mart{\'\i}n-Mart{\'\i}n, Linxi Fan,
  Guanzhi Wang, Shyamal Buch, Claudia D'Arpino, Sanjana Srivastava, Lyne~P
  Tchapmi, et~al.
\newblock igibson, a simulation environment for interactive tasks in large
  realisticscenes.
\newblock {\em arXiv preprint arXiv:2012.02924}, 2020.

\bibitem{singh2018learning}
Amanpreet Singh, Tushar Jain, and Sainbayar Sukhbaatar.
\newblock Learning when to communicate at scale in multiagent cooperative and
  competitive tasks.
\newblock In {\em ICLR}, 2020.

\bibitem{song2017semantic}
Shuran Song, Fisher Yu, Andy Zeng, Angel~X Chang, Manolis Savva, and Thomas
  Funkhouser.
\newblock Semantic scene completion from a single depth image.
\newblock In {\em CVPR}, 2017.

\bibitem{stergiopoulos2015distributed}
Yiannis Stergiopoulos, Michalis Thanou, and Anthony Tzes.
\newblock Distributed collaborative coverage-control schemes for non-convex
  domains.
\newblock {\em IEEE Transactions on Automatic Control}, 60(9):2422--2427, 2015.

\bibitem{stone2000multiagent}
Peter Stone and Manuela Veloso.
\newblock Multiagent systems: A survey from a machine learning perspective.
\newblock {\em Autonomous Robots}, 8(3):345--383, 2000.

\bibitem{striegnitz2011report}
Kristina Striegnitz, Alexandre Denis, Andrew Gargett, Konstantina Garoufi,
  Alexander Koller, and Mari{\"e}t Theune.
\newblock Report on the second challenge on generating instructions in virtual
  environments (give-2.5).
\newblock In {\em European Workshop on Natural Language Generation}, 2011.

\bibitem{sukhbaatar2016learning}
Sainbayar Sukhbaatar, Arthur Szlam, and Rob Fergus.
\newblock Learning multiagent communication with backpropagation.
\newblock {\em arXiv preprint arXiv:1605.07736}, 2016.

\bibitem{suttle2019multi}
Wesley Suttle, Zhuoran Yang, Kaiqing Zhang, Zhaoran Wang, Tamer Basar, and Ji
  Liu.
\newblock A multi-agent off-policy actor-critic algorithm for distributed
  reinforcement learning.
\newblock {\em arXiv preprint arXiv:1903.06372}, 2019.

\bibitem{tan1993multi}
Ming Tan.
\newblock Multi-agent reinforcement learning: Independent vs. cooperative
  agents.
\newblock In {\em ICML}, 1993.

\bibitem{thrun2002probabilistic}
Sebastian Thrun.
\newblock Probabilistic robotics.
\newblock {\em Communications of the ACM}, 45(3):52--57, 2002.

\bibitem{Wang_2021_CVPR}
Hanqing Wang, Wenguan Wang, Wei Liang, Caiming Xiong, and Jianbing Shen.
\newblock Structured scene memory for vision-language navigation.
\newblock In {\em CVPR}, 2021.

\bibitem{wang2020active}
Hanqing Wang, Wenguan Wang, Tianmin Shu, Wei Liang, and Jianbing Shen.
\newblock Active visual information gathering for vision-language navigation.
\newblock In {\em ECCV}, 2020.

\bibitem{williams1992simple}
Ronald~J Williams.
\newblock Simple statistical gradient-following algorithms for connectionist
  reinforcement learning.
\newblock {\em Machine learning}, 1992.

\bibitem{wu2018building}
Yi Wu, Yuxin Wu, Georgia Gkioxari, and Yuandong Tian.
\newblock Building generalizable agents with a realistic and rich 3d
  environment.
\newblock {\em arXiv preprint arXiv:1801.02209}, 2018.

\bibitem{xia2020interactive}
Fei Xia, William~B Shen, Chengshu Li, Priya Kasimbeg, Micael~Edmond Tchapmi,
  Alexander Toshev, Roberto Mart{\'\i}n-Mart{\'\i}n, and Silvio Savarese.
\newblock Interactive gibson benchmark: A benchmark for interactive navigation
  in cluttered environments.
\newblock {\em IEEE Robotics and Automation Letters}, 5(2):713--720, 2020.

\bibitem{xia2018gibson}
Fei Xia, Amir~R Zamir, Zhiyang He, Alexander Sax, Jitendra Malik, and Silvio
  Savarese.
\newblock Gibson env: Real-world perception for embodied agents.
\newblock In {\em CVPR}, 2018.

\bibitem{yu2021surprising}
Chao Yu, Akash Velu, Eugene Vinitsky, Yu Wang, Alexandre Bayen, and Yi Wu.
\newblock The surprising effectiveness of mappo in cooperative, multi-agent
  games, 2021.

\bibitem{zhang2017neural}
Jingwei Zhang, Lei Tai, Joschka Boedecker, Wolfram Burgard, and Ming Liu.
\newblock Neural slam: Learning to explore with external memory.
\newblock {\em arXiv preprint arXiv:1706.09520}, 2017.

\bibitem{zhang2019multi}
Kaiqing Zhang, Zhuoran Yang, and Tamer Ba{\c{s}}ar.
\newblock Multi-agent reinforcement learning: A selective overview of theories
  and algorithms.
\newblock {\em arXiv preprint arXiv:1911.10635}, 2019.

\bibitem{zhang2018fully}
Kaiqing Zhang, Zhuoran Yang, Han Liu, Tong Zhang, and Tamer Basar.
\newblock Fully decentralized multi-agent reinforcement learning with networked
  agents.
\newblock In {\em ICML}, 2018.

\bibitem{zhang2020multi}
Tianjun Zhang, Huazhe Xu, Xiaolong Wang, Yi Wu, Kurt Keutzer, Joseph~E
  Gonzalez, and Yuandong Tian.
\newblock Multi-agent collaboration via reward attribution decomposition.
\newblock {\em arXiv preprint arXiv:2010.08531}, 2020.

\bibitem{zheng2017magent}
Lianmin Zheng, Jiacheng Yang, Han Cai, Ming Zhou, Weinan Zhang, Jun Wang, and
  Yong Yu.
\newblock Magent: A many-agent reinforcement learning platform for artificial
  collective intelligence.
\newblock In {\em AAAI}, 2018.

\bibitem{zhu2016targetdriven}
Yuke Zhu, Roozbeh Mottaghi, Eric Kolve, Joseph~J Lim, Abhinav Gupta, Li
  Fei-Fei, and Ali Farhadi.
\newblock Target-driven visual navigation in indoor scenes using deep
  reinforcement learning.
\newblock In {\em ICRA}, 2017.

\end{thebibliography}
}

\clearpage

\appendix
In the supplementary material, we first elaborate on our dataset construction (\S\ref{sec:dataset_construct}). Then, more implementation details and discussions on evaluation metric designs are provided in \S\ref{sec:id} and \S\ref{sec:eva}. Finally, we present some representative visual results in \S\ref{sec:vs}.

\section{Details of CollaVN} \label{sec:dataset_construct}
Our CollaVN dataset has three subdatasets, corresponding to the three MAVN tasks: CommonGoal, SpecificGoal and Ad-HoCoop.

\noindent \textbf{CommonGoal-CollaVN}: It contains three subsets, which are created for different agent numbers (\ie, $N\!=\!\{2,3,4\}\!$), and agents in each episode are assigned a same target image. For each subset of different agent number, we randomly sample 5K episodes per training scene. For each scene in val/test split, we generate 0.5K/1K episodes per configuration, \ie, $\{$\textit{easy} (1.5-3 m),  {\textit{medium} (3-5 m)}, {\textit{hard} (5-10 m)}$\}$. The distance range means that the initial start and target position of all agents are within this limit. Finally, each subset has 125K/7.5K/15K samples for \texttt{train}/\texttt{val}/\texttt{test}, \ie, $25\!\times\!5\text{K}/3\!\times\!5\!\times\!0.5\text{K}/3\!\times\!5\!\times\!1\text{K}\!=\!125\text{K}/7.5\text{K}/15\text{K}$.
Detailed statistics of CommonGoal-CollaVN subdataset are shown in Table \ref{tab:com}.

\noindent \textbf{SpecificGoal-CollaVN}: Three subsets are created for different agent numbers (\ie, $N\!=\!\{2,3,4\}\!$), but agents in each episode are assigned different target images.
Similarly, each subset has 125K/7.5K/15K samples for \texttt{train}/\texttt{val}/ \texttt{test}.
Detailed statistics of SpecificGoal-CollaVN subdataset are shown in Table \ref{tab:specific}.

\noindent \textbf{Ad-Hoc-CollaVN}: Two subsets are created for different ad-hoc team size changing situations (\ie, $N\!\!:\!2\!\rightarrow\!3$ and $N\!\!:\!3\!\rightarrow\!2$), and each has 125K/7.5K/15K samples for \texttt{train}/\texttt{val}/\texttt{test}. Agents in each episode are assigned a same target image.  For $N\!\!:\!2\!\rightarrow\!3$, its \texttt{train} set is the one in  \textit{CommonGoal}-CollaVN with $N\!=\!2$, but its \texttt{val} and \texttt{test} sets are new generated. Each training sample in $N\!=\!2$ contains 2 agents while each val/test sample has 3 agents. The subset of $N\!\!:\!3\!\rightarrow\!2$ is built in a similar way. Detailed statistics of Ad-Hoc-CollaVN subdataset are shown in Table \ref{tab:ad}.

\begin{table}
    \centering
    \small
    \resizebox{0.35\textwidth}{!}{
        \setlength\tabcolsep{6pt}
		\renewcommand\arraystretch{1.0}
        \begin{tabular}{c|l||ccc}
            \hline\thickhline
            \rowcolor{mygray}
& &\multicolumn{3}{c}{CommonGoal} \\
\cline{3-5}
\cline{3-5}
\cline{3-5}
\rowcolor{mygray}
\multirow{-2}{*}{$N$} &\multirow{-2}{*}{Difficulty} & Train & Val & Test \\
\hline\hline
\multirow{3}{*}{2}&Easy (1.5-3m)     & 125K  & 2.5K & 5K    \\
&Medium (3-5m)  & 125K  & 2.5K & 5K    \\
&Hard (5-10m) & 125K  & 2.5K & 5K    \\\hline\hline

\multirow{3}{*}{3}&Easy (1.5-3m)    & 125K   & 2.5K & 5K    \\
&Medium (3-5m)  & 125K  & 2.5K & 5K    \\
&Hard (5-10m) & 125K   & 2.5K & 5K  \\\hline\hline

\multirow{3}{*}{4}&Easy (1.5-3m)    & 125K   & 2.5K & 5K \\
&Medium (3-5m)  & 125K  & 2.5K & 5K  \\
&Hard (5-10m) & 125K   & 2.5K & 5K \\
\hline
\end{tabular}
}
\captionsetup{font=small}
	\caption{\small \textbf{Dataset Statistics} on CommonGoal-CollaVN. }
	\label{tab:com}
\end{table}

\begin{table}
    \centering
    \small
    \resizebox{0.35\textwidth}{!}{
        \setlength\tabcolsep{6pt}
		\renewcommand\arraystretch{1.0}
        \begin{tabular}{c|l||ccc}
            \hline\thickhline
            \rowcolor{mygray}
& &\multicolumn{3}{c}{SpecificGoal} \\
\cline{3-5}
\cline{3-5}
\cline{3-5}
\rowcolor{mygray}
\multirow{-2}{*}{$N$} &\multirow{-2}{*}{Difficulty} & Train & Val & Test \\
\hline\hline
\multirow{3}{*}{2}&Easy (1.5-3m)     & 125K  & 2.5K & 5K    \\
&Medium (3-5m)  & 125K  & 2.5K & 5K    \\
&Hard (5-10m) &  125K  & 2.5K & 5K    \\\hline\hline

\multirow{3}{*}{3}&Easy (1.5-3m)    & 125K   & 2.5K & 5K    \\
&Medium (3-5m)  & 125K  & 2.5K & 5K    \\
&Hard (5-10m) & 125K   & 2.5K & 5K  \\\hline\hline

\multirow{3}{*}{4}&Easy (1.5-3m)    & 125K   & 2.5K & 5K \\
&Medium (3-5m)  & 125K  & 2.5K & 5K  \\
&Hard (5-10m) & 125K   & 2.5K & 5K \\
\hline
\end{tabular}
}
\captionsetup{font=small}
	\caption{\small \textbf{Dataset Statistics} on SpecificGoal-CollaVN. }
	\label{tab:specific}
\end{table}

\begin{table}
    \centering
    \small
    \resizebox{0.5\textwidth}{!}{
        \setlength\tabcolsep{6pt}
		\renewcommand\arraystretch{1.0}
        \begin{tabular}{c|l||ccc}
            \hline\thickhline
            \rowcolor{mygray}
& &\multicolumn{3}{c}{Ad-HoCoop} \\
\cline{3-5}
\cline{3-5}
\cline{3-5}
\rowcolor{mygray}
\multirow{-2}{*}{$N$} &\multirow{-2}{*}{Difficulty} & Train & Val & Test \\
\hline\hline
\multirow{3}{*}{Add ($N:\!2\!\rightarrow\!3$)}&Easy(1.5-3m)    & 125K   & 2.5K & 5K \\
&Medium(3-5m) & 125K   & 2.5K & 5K \\
&Hard(5-10m) & 125K   & 2.5K & 5K \\\hline\hline

\multirow{3}{*}{Remove ($N:\!3\!\rightarrow\!2$)}&Easy(1.5-3m)    & 125K   & 2.5K & 5K \\
&Medium(3-5m) & 125K   & 2.5K & 5K \\
&Hard(5-10m) & 125K   & 2.5K & 5K \\

\hline
\end{tabular}
}
\captionsetup{font=small}
	\caption{\small \textbf{Dataset Statistics} on Ad-Hoc-CollaVN. }
	\label{tab:ad}
\end{table}

\section{Implementation Details} \label{sec:id}
\subsection{Reward Structure} \label{sec:reward}
During training, rewards are provided to all agents individually at every step. The reward includes two terms: \textbf{i)} $+1.00\!\times\!\triangle d$, where $d$ is the geodesic distance between agent's current location and target location, and $\triangle d$ refers to the distance change after adapting a navigation action at this step. $+1.00$ is the weight of goal-driven reward. \textbf{ii)} A penalty of $-0.05$ whenever the agent collides with environment or other agents. The maximum total reward achievable for a single agent is $+D$, where $D$ is the initial distance between agent and goal,  \ie, reaching the goal through the shortest path without any collision. Our models are trained to maximize the expected discounted cumulative gain with a discounting factor $\gamma=0.99$.

\begin{figure*}[t]
  \centering
      \includegraphics[width=0.99\linewidth]{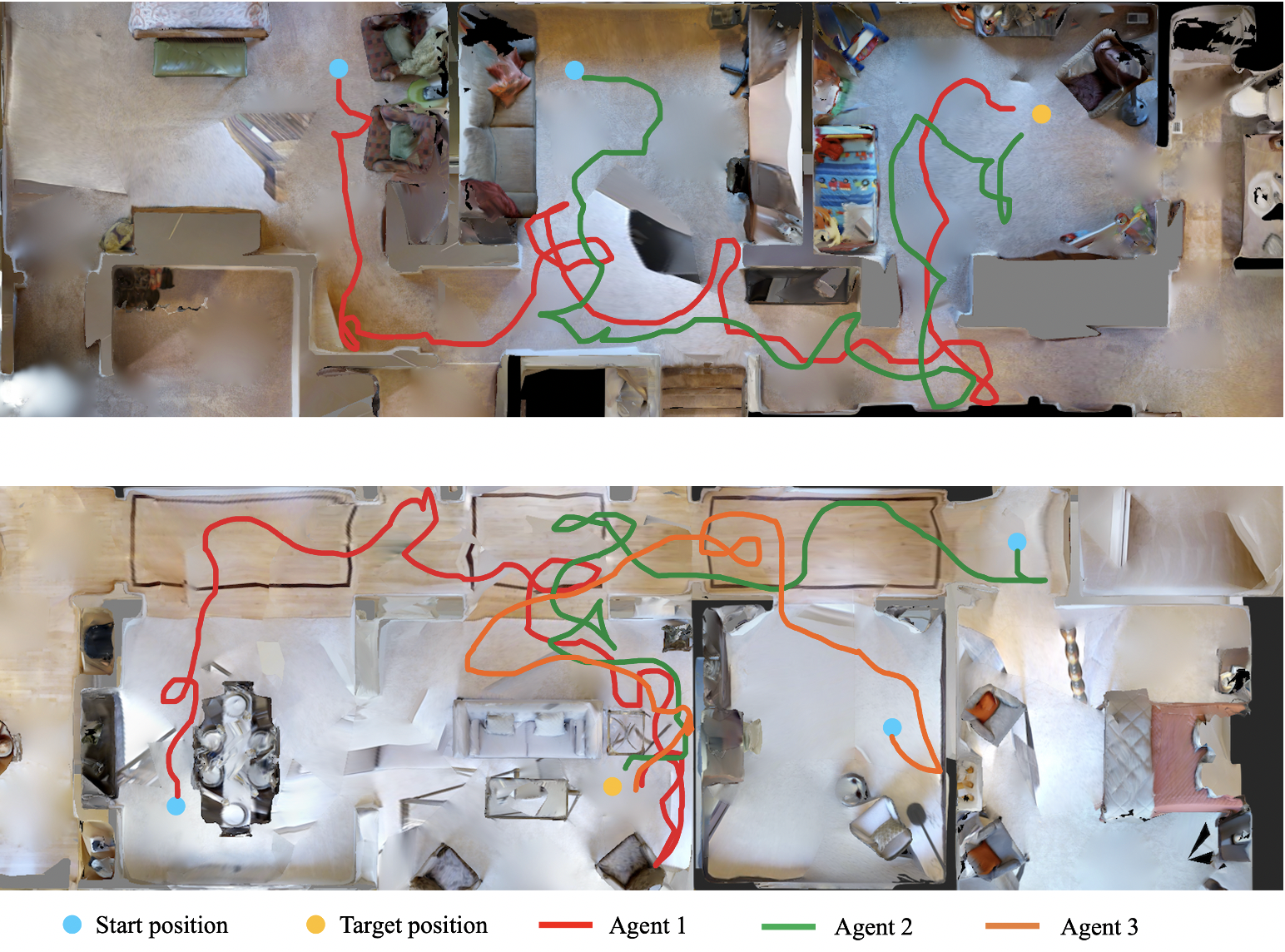}
\captionsetup{font=small}
	\caption{\small{Trajectory of two or three agents on CommonGoal task (\textbf{Top}: two agents ~\textbf{Down}: three agents).
}
}
\label{fig:commogoal}
\end{figure*}

\begin{table}
    \centering
    \small
    \resizebox{0.50\textwidth}{!}{
        \setlength\tabcolsep{18pt}
		\renewcommand\arraystretch{1.0}
        \begin{tabular}{c||ccc}
            \hline\thickhline
            \rowcolor{mygray}

\rowcolor{mygray}
Indicator & 2 (3-5m) & 3 (3-5m) & 4 (3-5m) \\
\hline\hline
SR    & 16.55  & 17.50 & 18.40    \\
SPL    & 0.08  & 0.08 & 0.09    \\
SSR    & 0.41  & 0.45 & 0.48 \\
\hline
\end{tabular}
}
\captionsetup{font=small}
	\caption{\small \textbf{SPL \textit{vs} SSR.} Performance of our model on \textit{CommonGoal} task with medium difficulty and different team sizes ($N\!=\!2, 3, 4$). }
	\label{tab:indicator}
\end{table}

\begin{figure*}[t]
  \centering
      \includegraphics[width=0.99\linewidth]{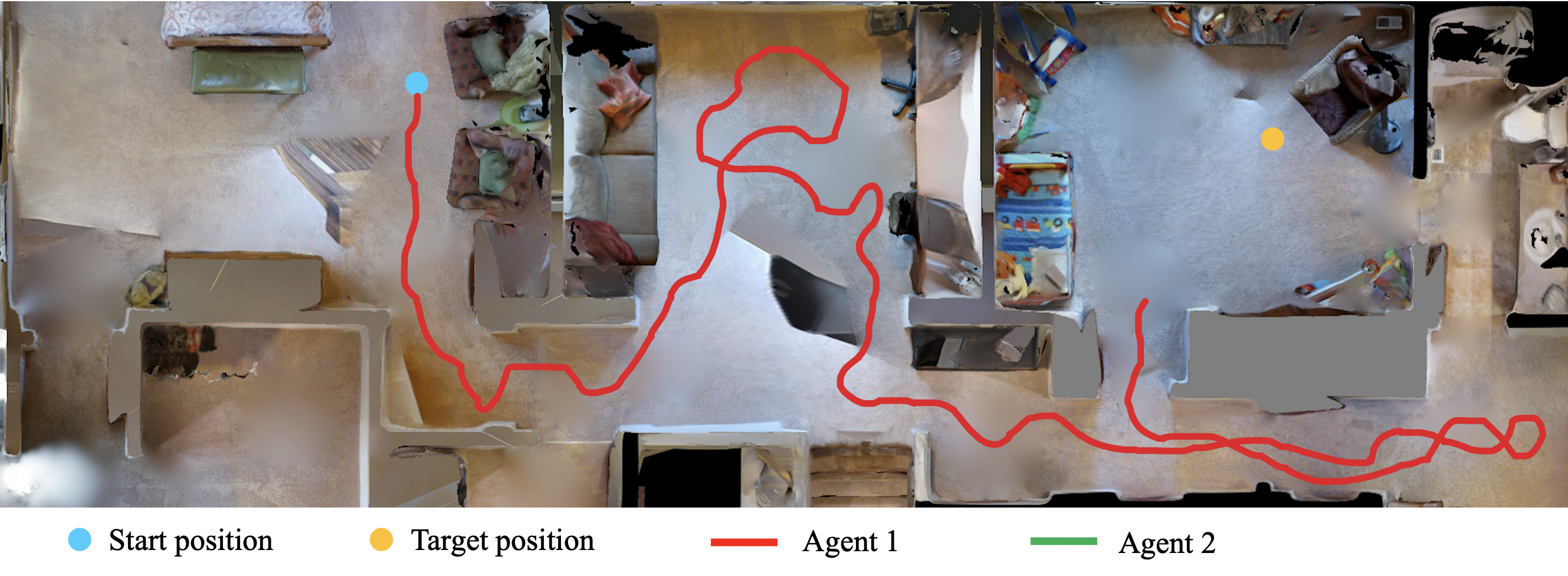}
\captionsetup{font=small}
	\caption{\small{Trajectory of single agent (same initialization with two agents of CommonGoal task).
}
}
\label{fig:single}
\end{figure*}

\begin{figure*}[t]
  \centering
      \includegraphics[width=0.99\linewidth]{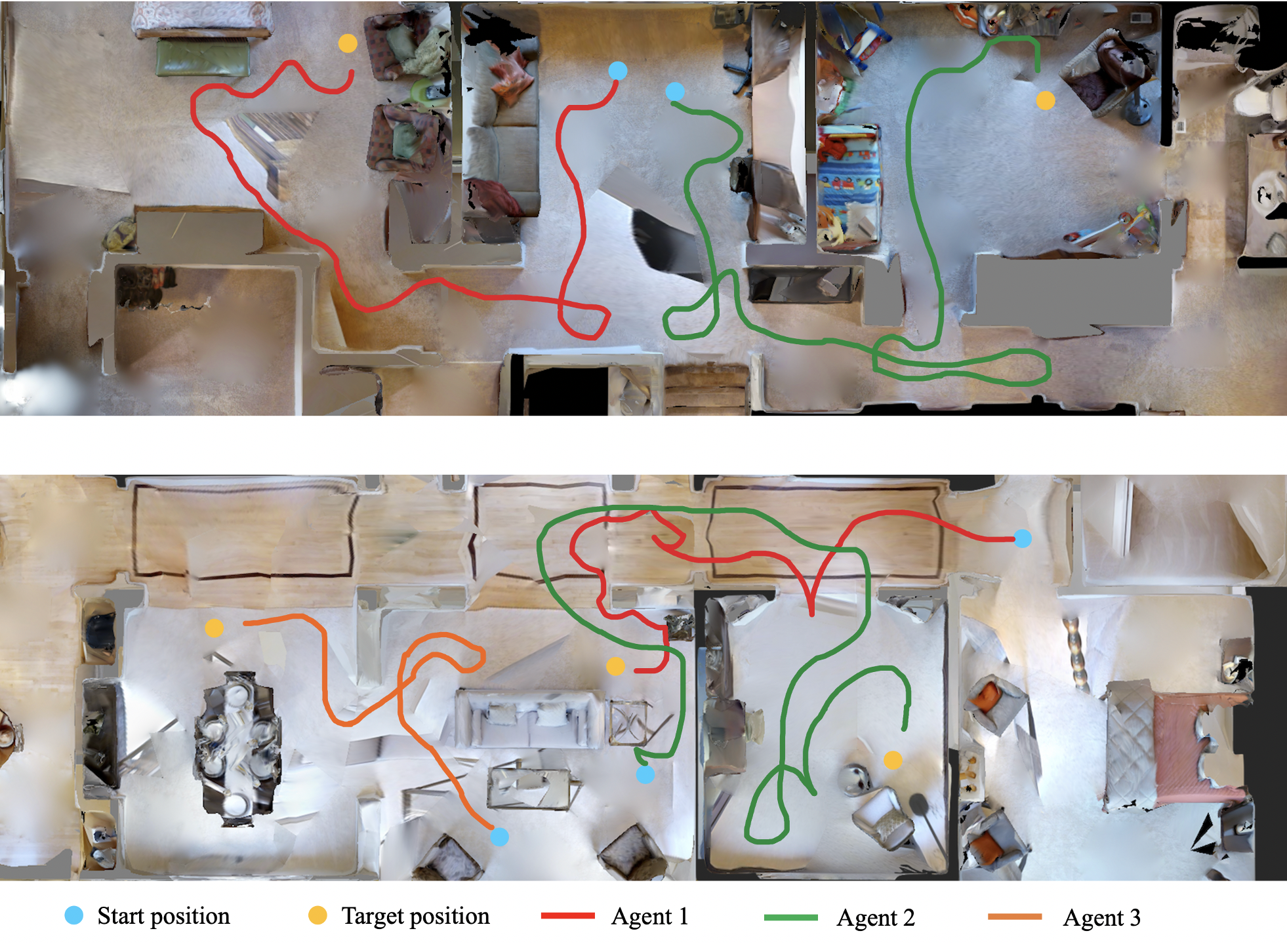}
\captionsetup{font=small}
	\caption{\small{Trajectory of two or three agents on Specific task (\textbf{Top}: two agents ~\textbf{Down}: three agents). }
}
\label{fig:specificgoal}
\end{figure*}

\begin{figure*}[t]
  \centering
      \includegraphics[width=0.99\linewidth]{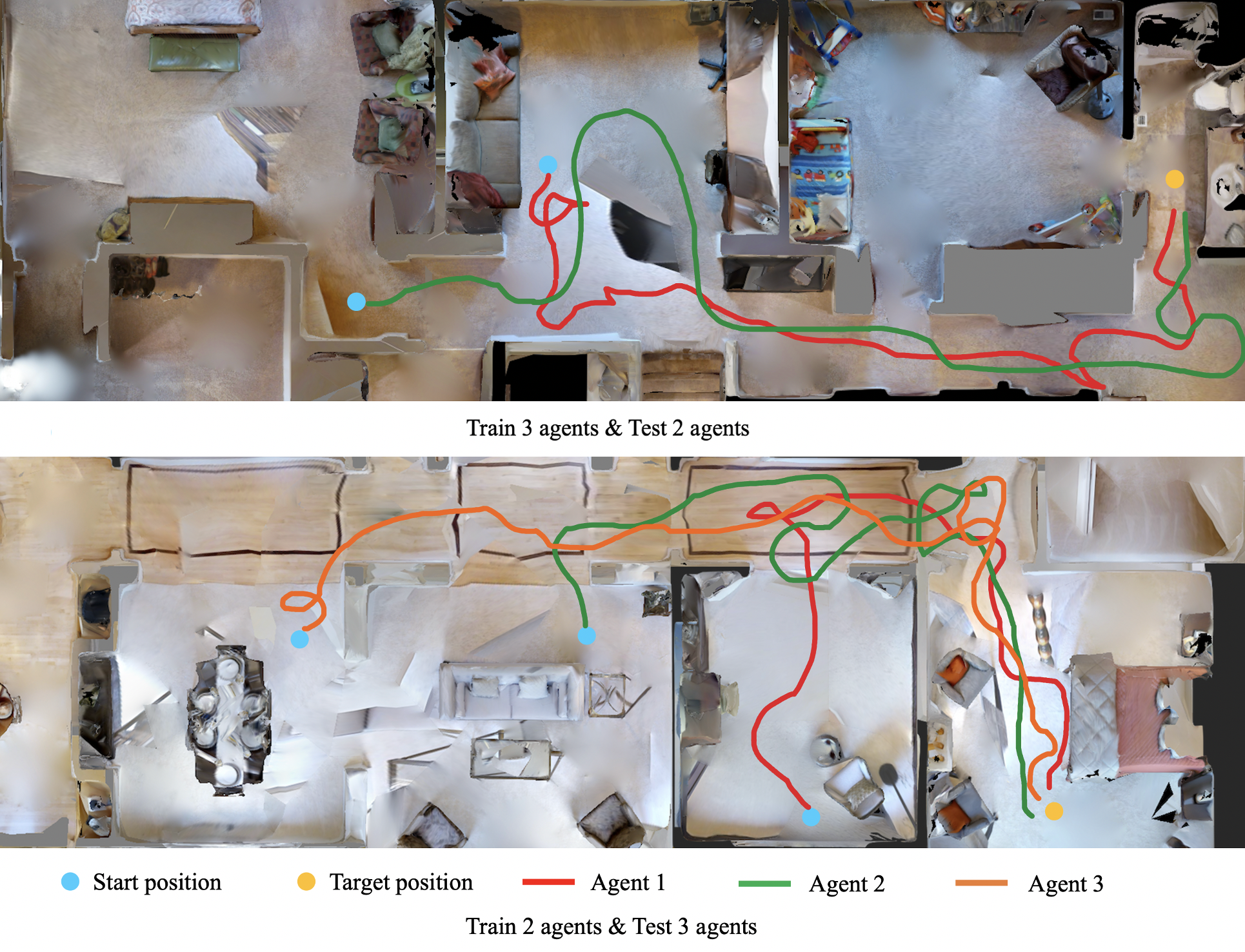}
\captionsetup{font=small}
	\caption{\small{Trajectory of add and remove on Ad-HoCoop task (\textbf{Top}: remove ($N:3\!\rightarrow\!2$) ~\textbf{Down}: add ($N:2\!\rightarrow\!3$)).}
}
\label{fig:ad_hoc}
\end{figure*}

\subsection{Hyperparameter Details} \label{sec:hyper}
We use PyTorch for implementing and training
our model. As for map building, we follow~\cite{chaplot2020learning}, maintain a FIFO memory of size 5000 for training. After one step in each thread, we perform 10 updates to the map building module with a batch size of 64.

During training, we train our agents with 25 parallel threads, with each thread using one scene in the training set. We use Proximal Policy Optimization (PPO)~\cite{schulman2017proximal}, with 5 mini-batches, and 8 epochs in each PPO update. Our PPO implementation is based on ~\cite{pytorchrl}. We use Adam optimizer with a learning rate of 0.00001, a discount factor of $\gamma$ = 0.99, an entropy coefficient of 0.001, value loss coefficient of 0.5 for training.

\section{Evaluation Details} \label{sec:eva}
\noindent \textbf{Why SSR instead of SPL?} SPL was introduced in~\cite{anderson2018evaluation}:
\begin{equation}
	SPL = \frac{1}{NK}\sum_{k=1}^{K}\sum_{n=1}^{N}\tau_{k,n}\frac{l^{g}_{k,n}}{\text{max}(l^{g}_{k,n}, ~l_{k,n})}.
\end{equation}
where $N$ is the agent number, $K$ equals the number of test episodes, $\tau_{k,n}\!\in\!\{0,1\}$ is a binary indicator of success in episode $k$ for agent $n$, $l^g_{k,n}$ is the geodesic distance from starting position to the goal, and $l_{k,n}$ the length of the path actually taken by agent.

The value of ${l^g_{k,n}}/{\text{max}(l^g_{k,n}, ~l_{k,n})}$ is always less than one ($<$1). If we perform a long-term or difficult navigation task, the success rate is usually low and the navigation path is very long. In this case, SPL will be very small and lose discriminability. This phenomenon can be observed in Table~\ref{tab:indicator}, which shows the performance of our model on CommonGoal task with medium difficulty and different team sizes. SPL basically stays still with the success rate changing.

To resolve this problem, we introduce SSR (Success weighted by Step Ratio):
\begin{equation}
	SSR = \frac{1}{NK}\sum_{k=1}^{K}\sum_{n=1}^{N}\tau_{k,n}\frac{T}{\text{min}(T, ~T_{k,n})}.
\end{equation}
$T$ denotes the allowed maximum time step and $T_{k,n}$ the number of navigation steps used by agent $n$ in episode $k$. As shown in Table~\ref{tab:indicator}, SSR is more discriminative than SPL.

\section{Qualitative Results}\label{sec:vs}
In this section, we provide more intuitive results of our model.\\
\subsection{Trajectory Result}
The visualization of navigation trajectories of three task settings with different team size ($N \!=\! 2, 3$) are shown in Fig.~\ref{fig:commogoal}, Fig.~\ref{fig:specificgoal}, and Fig.~\ref{fig:ad_hoc}.  The start and target locations are represented by blue circles and orange circles, respectively. From these figures, we can observe that the agents trained with our method is able to learn a robust policy to collaboratively explore the environment and successfully reaches the target location. From Fig.~\ref{fig:ad_hoc}, we find that our method can adapt to different team sizes at test times well.
\subsection{Effect of Communication}
To show the effectiveness of communication in MVAN task, we demonstrate the trajectory of a single agent system in Fig.~\ref{fig:single}. We compare the effect by deploying this agents for a particular initialization of an episode, \ie,  the
scene, agents' start location and target location are same as CommonGoal task with team size of two  (Top figure in Fig.~\ref{fig:commogoal}). As shown in Fig.~\ref{fig:single}, the single agent failed to reach the target position within maximum time step (it spends too much time on exploring a wrong room). However, in CommonGoal task, we find that implicit communication can allow agents to avoid this wrong exploration as much as possible based on the experience of other agents and achieve the task faster.

\end{document}